%% file: arxiv.tex
\pdfoutput=1

\documentclass[11pt]{article}

\usepackage[final]{acl}

\input{preamble}
\usepackage{times}
\usepackage{latexsym}

\usepackage[T1]{fontenc}

\usepackage[utf8]{inputenc}

\usepackage{microtype}

\usepackage{inconsolata}

\usepackage{graphicx}
\usepackage{epigraph}
\usepackage{csquotes}
\usepackage[justification=justified]{caption}

\usepackage{xcolor}
\usepackage{epigraph}
\input{color}
\input{def}

\newcommand{\bdicon}{\faIcon{paw}}
\newcommand{\unvicon}{\faIcon{university}}
\newcommand{\shieldicon}{\faIcon{shield-alt}}

\renewcommand{\arraystretch}{1.05}

\title{The Mirage of Model Editing: Revisiting Evaluation in the \textsc{\scalebox{1.05}Wild}\\
\normalsize{\textit{\textcolor{flamingo}{Are We Really Making Much Progress?}}}}  %

\author{Wanli Yang\textsuperscript{\tiny\shieldicon\unvicon}\hspace{2em} Fei Sun\textsuperscript{\tiny\shieldicon ~\textcolor{matisse}{\faIcon[regular]{envelope}}} \\ %
  {\bf Jiajun Tan}\textsuperscript{\tiny\shieldicon\unvicon} \hspace{0.6em} \textbf{Xinyu Ma}$\textsuperscript{\tiny\bdicon}$ \hspace{0.6em} \textbf{Qi Cao}\textsuperscript{\tiny\shieldicon} \hspace{0.6em}  \textbf{Dawei Yin}$\textsuperscript{\tiny\bdicon}$ \hspace{0.6em} \textbf{Huawei Shen}\textsuperscript{\tiny\shieldicon\unvicon} \hspace{0.6em} \textbf{Xueqi Cheng}\textsuperscript{\tiny\shieldicon\unvicon}\\
  \textsuperscript{\tiny\shieldicon}State Key Laboratory of AI Safety, Institute of Computing Technology, CAS\\
  $\textsuperscript{\tiny\unvicon}$University of Chinese Academy of Sciences  \hspace{2.1em} $\textsuperscript{\tiny\bdicon}$Baidu Inc. \\ %
 \tt{yangwanli24z@ict.ac.cn} \,\,\,\,\,
 \textsuperscript{\tiny\textcolor{matisse}{\faIcon[regular]{envelope}}}\tt{sunfei@ict.ac.cn}
}

\begin{document}
\maketitle

\renewcommand*{\thefootnote}{\tiny\textcolor{matisse}{\faIcon[regular]{envelope}}}
\footnotetext{Corresponding author: Fei Sun (\href{sunfei@ict.ac.cn}{sunfei@ict.ac.cn})}
\renewcommand*{\thefootnote}{\arabic{footnote}}

\begin{abstract}

Despite near-perfect results reported in the literature, the effectiveness of model editing in real-world applications remains unclear. %
To bridge this gap, we introduce QAEdit, a new benchmark aligned with widely used question answering (QA) datasets, and \textsc{Wild}, a task-agnostic evaluation framework designed to better reflect real-world usage of model editing.
Our single editing experiments show that current editing methods perform substantially worse than previously reported (38.5\% vs. 96.8\%).
We demonstrate that it stems from issues in the synthetic evaluation practices of prior work.
Among them, the most severe is the use of \textit{teacher forcing} during testing, which leaks both content and length of the ground truth, leading to overestimated performance.
Furthermore, we simulate practical deployment by sequential editing, revealing that current approaches fail drastically with only 1000 edits.
This work calls for a shift in model editing research toward rigorous evaluation and the development of robust, scalable methods that can reliably update knowledge in LLMs for real-world use\footnote{Code and data are released at \url{https://github.com/WanliYoung/Revisit-Editing-Evaluation}.}.

\end{abstract}

\input{Section/intro_new}

\input{Section/2_Related_Works}

\input{Section/3_Benchmark}

\input{Section/4_Evaluation}

\input{Section/5_SingleEdit}

\input{Section/6_MetricsAnalysis}

\input{Section/7_SequentialEdit}

\input{Section/8_Conclusion}

\input{Section/9_Limitations}

\section*{Acknowledgments}

This work was supported by the National Key
R\&D Program of China (2022YFB3103700, 2022YFB3103704), the Strategic Priority Research Program of the Chinese Academy of Sciences (XDB0680201), the Beijing Natural Science Foundation (4252023), and the Innovation Funding of ICT, CAS (E361120).
We would also like to thank Wenyue Hua for her valuable discussions and insightful suggestions.

\newpage
\bibliography{custom}

\newpage

\input{Section/Appendix}

\end{document}

%% file: preamble.tex
\usepackage{booktabs}
\usepackage{multirow}
\usepackage[detect-weight, mode=text]{siunitx}
\usepackage[inline]{enumitem}
\usepackage{subcaption}

\usepackage{array}

\usepackage{bm}
\usepackage{adjustbox}
\usepackage{tcolorbox}
\usepackage{colortbl}
\usepackage[frozencache,cachedir=minted-cache]{minted}
\usepackage{svg}

\usepackage{hyperref}
\makeatletter
\def\UrlAlphabet{%
      \do\a\do\b\do\c\do\d\do\e\do\f\do\g\do\h\do\i\do\j%
      \do\k\do\l\do\m\do\n\do\o\do\p\do\q\do\r\do\s\do\t%
      \do\u\do\v\do\w\do\x\do\y\do\z\do\A\do\B\do\C\do\D%
      \do\E\do\F\do\G\do\H\do\I\do\J\do\K\do\L\do\M\do\N%
      \do\O\do\P\do\Q\do\R\do\S\do\T\do\U\do\V\do\W\do\X%
      \do\Y\do\Z}
\def\UrlDigits{\do\1\do\2\do\3\do\4\do\5\do\6\do\7\do\8\do\9\do\0}
\g@addto@macro{\UrlBreaks}{\UrlOrds}
\g@addto@macro{\UrlBreaks}{\UrlAlphabet}
\g@addto@macro{\UrlBreaks}{\UrlDigits}
\makeatother

\usepackage{pifont}
\usepackage{marvosym}
\usepackage{fontawesome5}
\usepackage{simpleicons}

\usepackage{tikz}
\usetikzlibrary{shapes.symbols}
\usetikzlibrary{arrows.meta}
\usetikzlibrary{backgrounds}
\usetikzlibrary{positioning, fit, mindmap, trees, calc, tikzmark, shapes}
\usetikzlibrary{shapes.arrows, fadings, automata, tikzmark, decorations.pathreplacing, patterns}
\usetikzlibrary{shapes.geometric}
\usetikzlibrary{intersections}
\usepackage{tkz-euclide}

 \usepackage{pgfplots}
\usepgfplotslibrary{groupplots}
\pgfplotsset{compat=1.18}

%% file: color.tex
\definecolor{mine_font}{RGB}{0, 128, 0}
\definecolor{tea_green}{RGB}{214, 234, 193}
\definecolor{hint_green}{RGB}{226,246,209}
\definecolor{Madang}{RGB}{190,235,159}
\definecolor{yellow_green}{RGB}{198,222,119}
\definecolor{link_water}{RGB}{221, 232, 250}
\definecolor{celestial_blue}{RGB}{52, 152, 219}
\definecolor{shakespeare}{RGB}{85, 154, 193}
\definecolor{buttermilk}{RGB}{255,242,174}
\definecolor{chardonnay}{RGB}{250,196,114}
\definecolor{rajah}{RGB}{253,180,98}
\definecolor{fog}{RGB}{213, 193, 234}
\definecolor{melon}{RGB}{254,191,181}
\definecolor{sundown}{RGB}{249, 180, 181}
\definecolor{mona_lisa}{RGB}{246,152,134}
\definecolor{salmon}{RGB}{242,131,107}
\definecolor{blue_x}{RGB}{142, 207, 201}
\definecolor{orange_x}{RGB}{255, 190, 122}

\definecolor{saltpan}{RGB}{238, 243, 232}
\definecolor{aqua_spring}{RGB}{232, 243, 232}
\definecolor{tea_green}{RGB}{214, 234, 193}
\definecolor{Madang}{RGB}{190,235,159}
\definecolor{fringy_flower}{RGB}{194, 234, 193}
\definecolor{aero_blue}{RGB}{193, 234, 213}
\definecolor{pixie_green}{RGB}{183,214,170}
\definecolor{french_pass}{RGB}{195,232,246}
\definecolor{ice_cold}{RGB}{169,232,220}
\definecolor{pale_turquoise}{RGB}{172,240,242}
\definecolor{cruise}{RGB}{179,226,205}
\definecolor{sail}{RGB}{163,205,235}
\definecolor{spindle}{RGB}{179,205,227}
\definecolor{link_water}{RGB}{221, 232, 250}
\definecolor{periwinkle}{RGB}{203,213,232}
\definecolor{zanah}{RGB}{220, 233, 213}
\definecolor{frostee}{RGB}{217, 231, 214}
\definecolor{opal}{RGB}{199, 221, 211}
\definecolor{jet_stream}{RGB}{188, 214, 210}
\definecolor{skeptic}{RGB}{153, 187, 167}
\definecolor{hint_green}{RGB}{226,246,209}
\definecolor{snow_flurry}{RGB}{230,245,201}
\definecolor{surf_crest}{RGB}{205,230,208}
\definecolor{yellow_green}{RGB}{198,222,119}
\definecolor{cream}{RGB}{255,255,204}
\definecolor{pale_prim}{RGB}{255,255,179}
\definecolor{spring_sun}{RGB}{242,243,195}
\definecolor{portafino}{RGB}{245,237,160}
\definecolor{buttermilk}{RGB}{255,242,174}
\definecolor{cream_brulee}{RGB}{255, 229, 151}
\definecolor{dairy_cream}{RGB}{254,226,189}
\definecolor{champagne}{RGB}{254,217,166}
\definecolor{chardonnay}{RGB}{250,196,114}
\definecolor{manhattan}{RGB}{226,180,125}
\definecolor{rajah}{RGB}{253,180,98}
\definecolor{early_dawn}{RGB}{252,243,218}
\definecolor{egg_shell}{RGB}{238, 234, 215}
\definecolor{selago}{RGB}{243, 232, 243}
\definecolor{quartz}{RGB}{219,223,238}
\definecolor{fog}{RGB}{213, 193, 234}
\definecolor{languid_lavender}{RGB}{222,203,228}
\definecolor{watusi}{RGB}{254,221,207}
\definecolor{coral_andy}{RGB}{243,204,205}
\definecolor{cosmos}{RGB}{248,209,210}
\definecolor{melon}{RGB}{254,191,181}
\definecolor{azalea}{RGB}{234, 193, 194}
\definecolor{beauty_bush}{RGB}{235, 185, 179}
\definecolor{sundown}{RGB}{249, 180, 181}
\definecolor{mona_lisa}{RGB}{246,152,134}
\definecolor{salmon}{RGB}{242,131,107}

\definecolor{summer_sky}{RGB}{58, 151, 233}
\definecolor{chateau_green}{RGB}{72, 179, 96}
\definecolor{matisse}{RGB}{25, 104, 167}
\definecolor{allports}{RGB}{31, 106, 125}
\definecolor{sun_shade}{RGB}{255, 144, 68}
\definecolor{flamingo}{RGB}{237, 88, 85}
\definecolor{studio}{RGB}{128, 91, 160}

\definecolor{maya_blue}{RGB}{102, 204, 255}
\definecolor{feijoa}{RGB}{178,223,138}
\definecolor{sushi}{RGB}{117, 168, 47}
\definecolor{norway}{RGB}{158, 194, 132}
\definecolor{japanese_laurel}{RGB}{53, 116, 40}
\definecolor{see_green}{RGB}{161,228,195}
\definecolor{monte_carlo}{RGB}{135,204,194}
\definecolor{granny_smith_apple}{RGB}{150,214,150}
\definecolor{moss_green}{RGB}{170,216,176}
\definecolor{chateau_green}{RGB}{72, 179, 96}
\definecolor{opal}{RGB}{164,207,190}
\definecolor{acapulco}{RGB}{117, 170, 148}
\definecolor{viridian}{RGB}{55, 137, 122}
\definecolor{amazon}{RGB}{56, 123, 84}
\definecolor{asparagus}{RGB}{123, 160, 91}
\definecolor{fruit_salad}{RGB}{91, 160, 94}
\definecolor{puerto_rico}{RGB}{72, 179, 150}
\definecolor{mountain_meadow}{RGB}{0, 163, 136}
\definecolor{matisse}{RGB}{25, 104, 167}
\definecolor{allports}{RGB}{31, 106, 125}
\definecolor{astral}{RGB}{55, 111, 137}
\definecolor{spring_leaves}{RGB}{46, 83, 117}
\definecolor{biscay}{RGB}{44, 62, 80}
\definecolor{midnight}{RGB}{0, 29, 50}
\definecolor{amethyst}{RGB}{153, 102, 204}
\definecolor{studio}{RGB}{128, 91, 160}
\definecolor{tapestry}{RGB}{194, 109, 132}
\definecolor{atomic_tangerine}{RGB}{255, 153, 102}
\definecolor{amber}{RGB}{255, 191, 0}
\definecolor{casablanca}{RGB}{244, 178, 84}
\definecolor{california}{RGB}{233, 140, 58}
\definecolor{tomato}{RGB}{255, 97, 56} 
\definecolor{alizarin}{RGB}{233, 58, 64}

\definecolor{linen}{RGB}{251, 239, 227}
\definecolor{double_pearl_lusta}{RGB}{253, 242, 208}
\definecolor{oasis}{RGB}{253, 242, 208}
\definecolor{milan}{RGB}{255, 254, 169}
\definecolor{texas}{RGB}{245, 232, 123}
\definecolor{maize}{RGB}{249, 212, 156}

\definecolor{turmeric}{RGB}{211, 178, 76}
\definecolor{saffron}{RGB}{249,193,62}
\definecolor{my_sin}{RGB}{255, 176, 59}
\definecolor{tree_poppy}{RGB}{246, 154, 27}
\definecolor{jaffa}{RGB}{240, 131, 58}
\definecolor{crusta}{RGB}{254, 127, 44}
\definecolor{tahiti_gold}{RGB}{223, 102, 36}
\definecolor{outrageous_orange}{RGB}{255, 100, 45}
\definecolor{safety_orange}{RGB}{254, 106, 0}

\definecolor{azalea}{RGB}{251, 196, 196}
\definecolor{oyster_pink}{RGB}{238,206,205} 
\definecolor{coral_candy}{RGB}{242,208,205} 
\definecolor{baby_pink}{RGB}{246, 194, 192}
\definecolor{petite_orchid}{RGB}{223, 157, 155}
\definecolor{apricot}{RGB}{241,140,122}
\definecolor{NY_pink}{RGB}{228,136,113}
\definecolor{carmine_pink}{RGB}{231, 76, 60}
\definecolor{deep_carmine_pink}{RGB}{236, 50, 67}

\definecolor{wewak}{RGB}{244, 143, 150}
\definecolor{light_coral}{RGB}{244, 127, 123}
\definecolor{bittersweet}{RGB}{255,111,105}
\definecolor{carnation}{RGB}{245, 80, 86}
\definecolor{flamingo}{RGB}{237, 88, 85}
\definecolor{sunset_orange}{RGB}{242,89,75}
\definecolor{ku_crimson}{RGB}{243, 0, 25}
\definecolor{amaranth}{RGB}{234,46,73}
\definecolor{valencia}{RGB}{214, 87, 70}
\definecolor{chilean_fire}{RGB}{215, 87, 44}
\definecolor{mexican_red}{RGB}{170, 41, 37}

\definecolor{napa}{RGB}{163, 154, 137}

\definecolor{athens_gray}{RGB}{236, 240, 241}
\definecolor{gallery}{RGB}{240,240,240}
\definecolor{mercury}{RGB}{230,230,230}
\definecolor{platinum}{RGB}{228,228,228}
\definecolor{silver}{RGB}{191,191,191}
\definecolor{aluminum}{RGB}{153,153,153}
\definecolor{ship_gray}{RGB}{77,77,77}
\definecolor{tuatara}{RGB}{67, 67, 67}

\definecolor{malibu}{RGB}{110, 180, 240}
\definecolor{celestial_blue}{RGB}{52, 152, 219}
\definecolor{curious_blue}{RGB}{41, 128, 185}
\definecolor{french_blue}{RGB}{0, 112, 182}
\definecolor{matisse}{RGB}{25, 104, 167}
\definecolor{shakespeare}{RGB}{85, 154, 193}
\definecolor{seagull}{RGB}{128,177,211}
\definecolor{jelly_bean}{RGB}{45, 126, 150}
\definecolor{venice_blue}{RGB}{87, 135, 105}
\definecolor{boston_blue}{RGB}{68, 147, 161}

\definecolor{turquoise}{RGB}{41,217,194}
\definecolor{java}{RGB}{2,190,196}
\definecolor{riptide}{RGB}{141,211,199}
\definecolor{mountain_meadow}{RGB}{0, 163, 136}
\definecolor{free_speech_aquamarine}{RGB}{0, 156, 114}

\definecolor{cosmic_latte}{RGB}{222, 247, 229}
\definecolor{chinook}{RGB}{163, 232, 178}
\definecolor{padua}{RGB}{121, 189, 143}
\definecolor{ocean_green}{RGB}{79, 176, 112}
\definecolor{pastel_green}{RGB}{107, 227, 135}
\definecolor{chateau_green}{RGB}{69, 191, 85}
\definecolor{RoyalBlue}{RGB}{69, 191, 85}
\definecolor{pigment_green}{RGB}{0, 175, 79}
\definecolor{fern}{RGB}{101,197,117}
\definecolor{killarney}{RGB}{56, 113, 66}

%% file: def.tex
\newcommand{\numedit}[1]{#1}

\newcommand{\numrwe}[2]{%
    \begin{tikzpicture}[baseline]
        \pgfmathparse{#1 > #2 ? 1 : 0}
        \ifnum\pgfmathresult=0
            \pgfmathsetmacro{\percentdiff}{min(130, 130*(#2-#1)/#2)}
            \pgfmathsetmacro{\intensity}{\percentdiff}
            \fill[monte_carlo!\intensity!white, rounded corners=1] (-0.6em, -0.3em) rectangle (2.6em, 1em);
        \fi
        \node[inner sep=0pt] at (1em, 0.7ex) {#1};
    \end{tikzpicture}%
}

\newcommand{\colorbarvertical}{%
    \begin{tikzpicture}
    \shade[%
           top color=white, 
           bottom color=monte_carlo!150, rounded corners=0.6] (0, 0) rectangle (0.3, 6);
        \foreach \y in {0,0.1,...,1} {
            \pgfmathsetmacro{\percentage}{\y*100}
            \draw[white, semithick, decorate, decoration={bumps, segment length=1mm, amplitude=0.6}] (0.25, 6-\y*6) -- (0.3, 6-\y*6);
            \draw[white, semithick] (0, 6-\y*6) -- (0.05, 6-\y*6);
            \node[right, ] at (0.25, 6-\y*6) {\footnotesize \pgfmathprintnumber[fixed,precision=0]{\percentage}};
        }
    \node[right] at (0.25, 0) {\footnotesize \pgfmathprintnumber[fixed,precision=0]{100}};
    \node[right] at (0.15, 6.5) {\small \textbf{\%}};
    \node[right] at (0.15, 7.7) {};
    \end{tikzpicture}%
}

%% file: Section/intro_new.tex
\section{Introduction}

\epigraph{``\textit{If you can’t measure it, you can’t improve it.}''}{--- Lord Kelvin}

\noindent Model editing \cite{yao-etal-2023-editing,wang2024knowledge} has attracted widespread attention for its promising vision: enabling efficient and precise updates to specific knowledge within pretrained Large Language Models (LLMs) without retraining from scratch. %
Recent advances report near-perfect results on corresponding benchmarks \cite{meng2023locating, wang2024wise}, suggesting substantial progress toward this goal.
However, these results often come from synthetic, oversimplified evaluation settings (e.g., identical prompts for editing and testing; more in \S\ref{sec:eval}) that may fail to capture real-world complexities.
This disparity raises a critical question: \textit{Can these promising results in the literature translate to practical applications?}

To address this question, we propose to study model editing in QA tasks, which provide clear evaluation criteria and broad applicability.
This adaptation involves two key components: a real-world dataset and realistic evaluation.
For dataset, we create \textbf{QAEdit}, a tailored dataset derived from three widely-used QA datasets, enabling editing methods to update LLMs with answers grounded in real-world tasks.
For evaluation, we propose \textsc{\textbf{Wild}} (\textbf{W}ithout \textbf{I}ntervention, \textbf{L}ive \textbf{D}ecoding), a task-agnostic evaluation framework that follows standard QA evaluation protocols \cite{eval-harness}, assessing editing methods via the performance of edited LLMs on their previously failed questions.

\input{Fig/intro_case}

Our initial study reveals that current advanced editing methods achieve only a \textbf{38.5}\% average success rate on QAEdit, significantly lower than the results reported in previous studies.
This raises a question: \textit{Does the performance decline stem from QAEdit's real-world complexity, or from the shift of synthetic to \textsc{Wild} evaluation?}

To enable rigorous analysis, starting with single editing experiments, we evaluate six representative methods across three leading LLMs on QAEdit and two established editing benchmarks, using both evaluation frameworks.
As illustrated in Figure~\ref{fig:intro}, switching from synthetic to \textsc{Wild} evaluation consistently leads to a significant performance decline across editing methods and datasets. %
This drama-tic performance gap raises two critical questions: \textit{What differences between these frameworks drive such disparity, and which one most accurately reflects editing effectiveness?}

To answer them, we carefully examine the setups for both synthetic and \textsc{Wild} evaluations.
From this, we abstract four key modules (\textit{input}, \textit{generation strategy}, \textit{output truncation}, and \textit{metric}) and analyze their variations through controlled experiments. 
The results expose four critical limitations in current synthetic evaluation in model editing:
\begin{enumerate}[leftmargin=14pt, itemsep=-3pt, topsep=2pt]%
\item[\textcolor{black}{\ding{182}}] \textbf{input module}: using identical prompts for editing and testing overlooks the variability and unpredictability in real-world queries;
\item[\textcolor{black}{\ding{183}}] \textbf{generation strategy}: teacher forcing, which feeds the ground truth as input during decoding, artificially beautifies results by disregarding potential errors in the model’s own outputs;
\item[\textcolor{black}{\ding{184}}] \textbf{output truncation}: using target answer length to truncate outputs conceals errors (e.g., repetition, irrelevant, or incorrect information) that would occur with natural stopping criteria;
\item[\textcolor{black}{\ding{185}}] \textbf{metric}: match ratio may inflate performance by rewarding partial matches of incorrect answers.
\end{enumerate} %
Among these issues, \textbf{teacher forcing} and \textbf{target length truncation} cause the most significant overestimation, as they rely on ground truth that is unavailable in real-world scenarios.
This highlights that \textbf{synthetic evaluation, reliant on such idealized or even unrealistic conditions, fails to accurately measure true editing effectiveness}.

After uncovering evaluation issues via single editing analysis, we return to our initial question: how do editing methods perform under realistic conditions? 
In practice, editing requests arrive continuously, making sequential editing a more genuine test of real-world applicability.
Under \textsc{Wild} evaluation, our sequential editing experiments show that current methods catastrophically fail to scale, with average success rates dropping to $\sim$\textbf{10}\% for only 1000 samples.

Our work, for the first time, exposes severe issues in current evaluation of model editing research and demonstrates substantial limitations of existing editing methods under real-world conditions. 
We hope this work will inspire more rigorous evaluation practices and motivate the development of algorithms that can truly fulfill the promise of model editing: to reliably and scalably update knowledge in LLMs \textit{for real-world applications}.

Our main contributions are as follows.
\begin{itemize}[itemsep=0pt, leftmargin=18pt, topsep=1pt, partopsep=1pt, parsep=1pt]
\item We introduce QAEdit, a benchmark tailored for real-world QA tasks, and establish a more rigorous evaluation framework, \textsc{Wild}.
\item We reveal that published model editing results are significantly inflated, and trace this overestimation to issues in synthetic evaluation practices, identified through modular analysis.
\item We expose the severe scalability challenges of current editing methods in practical applications through sequential editing experiments.
\end{itemize}

%% file: Fig/intro_case.tex
\pgfplotsset{
axis background/.style={fill=gallery!62},
grid=both,
  xtick pos=left,
  ytick pos=left,
  tick style={
    major grid style={style=white,line width=1pt},
    minor grid style=gallery!62,
    draw=none,
  },
  minor tick num=1,
}

\begin{figure}
\centering
\resizebox{\linewidth}{!}{
\begin{tikzpicture}
\begin{groupplot}[
    group style={group size=2 by 1,
        horizontal sep = 48pt,
        }, 
        width=1\linewidth,
        height=0.618\linewidth,
        enlarge x limits=0.15,
        xlabel=\textbf{\large ROME},
        ylabel=Success Rate (\%),
        ylabel shift={-5pt},
        ymin=40, ymax=102,
        xticklabels={ZsRE, \textsc{CounterFact}, \textbf{QAEdit}},
        xtick={1, 2, 3},
        ybar=5pt,%
        every axis plot/.style={bar width=15pt},
        ymajorgrids,
        major grid style={draw=white},
        y axis line style={opacity=0},
        tickwidth=0pt,
	]
    \nextgroupplot[
    legend style = {
          draw=none, 
          draw opacity=0,
          fill=none,
          column sep = 2pt, 
          /tikz/every even column/.append style={column sep=5mm},
          legend columns = -1, 
          legend to name = grouplegend
          },
    ]

     \addplot [draw=none, fill=monte_carlo!160]
        coordinates {
          (1, 96.4)
          (2, 99.6)
          (3, 95.5)};  \addlegendentry{synthetic evaluation}
        \addplot [draw=none, fill=my_sin!120]
        coordinates {
          (1, 74.1)
          (2, 83.6)
          (3, 58.5)}; \addlegendentry{\textsc{Wild} evaluation}

       \draw[-stealth, ku_crimson, very thick, line width=3pt] (axis cs:0.86,96.4) to [out=-40, in=100] (axis cs:1.14,74.1);
       \draw[-stealth, ku_crimson, very thick, line width=3pt] (axis cs:1.86,99.6) to [out=-30, in=120] (axis cs:2.14,83.6);
       \draw[-stealth, ku_crimson, very thick, line width=3pt] (axis cs:2.86,95.5) to [out=-55, in=90] (axis cs:3.14,58.5);
       
    \nextgroupplot[
    xlabel=\textbf{\large WISE},
    ymin=0, ymax=102,
    ]
     \addplot [draw=none, fill=monte_carlo!160]
        coordinates {
          (1, 99.9)
          (2, 99.9)
          (3, 99.8)};  %
        \addplot [draw=none, fill=my_sin!120]
        coordinates {
          (1, 13.9)
          (2, 52.1)
          (3, 21.6)}; %

         \draw[-stealth, ku_crimson, very thick, line width=3pt] (axis cs:0.86, 99.9) to [out=-50, in=90] (axis cs:1.14,13.9);
       \draw[-stealth, ku_crimson, very thick, line width=3pt] (axis cs:1.86, 99.9) to [out=-45, in=90] (axis cs:2.14,52.1);
       \draw[-stealth, ku_crimson, very thick, line width=3pt] (axis cs:2.86, 99.8) to [out=-60, in=90] (axis cs:3.14,21.6);
       
\end{groupplot}

\node at ($(group c1r1) + (100pt, 55pt)$) {\pgfplotslegendfromname{grouplegend}};

\end{tikzpicture}
}
\captionsetup{skip=3pt}
\caption{Comparison of synthetic and \textsc{Wild} evaluation for ROME and WISE on Llama-2-7b-chat.}
\label{fig:intro}
\end{figure}

%% file: Section/2_Related_Works.tex
\section{Related Works}

\subsection{Model Editing Methodologies}

Existing model editing methods can be categorized into the following four types:

\paragraph{Extension Based.}
These methods update LLMs by adding trainable parameters to encode new knowledge, e.g., additional neurons in FFN  \cite{dong-etal-2022-calibrating, huang2023transformerpatcher} or specialized memory modules \cite{hartvigsen2023aging, wang2024wise}, while preserving pretrained weights.

\paragraph{Fine-tuning Based.}
Fine-tuning offers a straightforward approach to update LLMs' knowledge but faces catastrophic forgetting.
Recent works mitigate this by constraining parameter changes \cite{zhu2020modifyingmemoriestransformermodels} or leveraging Parameter-Efficient Fine-Tuning (PEFT) \cite{han2024peft} to limit modification scope \cite{MELOaaai24,wang-etal-2024-roselora}.

\paragraph{Meta Learning.}
Employing meta learning, KE \cite{decao2021editing}, MEND \cite{mitchell2022fast}, and MALMEN \cite{tan23malmen} train hypernetworks to predict effective gradients or parameter alterations for knowledge integration. 

\paragraph{Locate-Then-Edit.}
Based on the investigation of knowledge mechanisms in LLMs \cite{geva-etal-2021-transformer, geva-etal-2022-transformer}, KN \cite{dai2022knowledge}, ROME \cite{meng2023locating}, and PMET \cite{aaai24pmet} utilize knowledge attribution and causal tracing to pinpoint target knowledge to specific parameters, then perform localized editing.
Furthermore, MEMIT \cite{meng2023massediting} and EMMET \cite{gupta-etal-2024-unified} extend this for massive editing in a batch.

\subsection{Evaluation of Model Editing}

Current evaluation of model editing primarily focuses on editing effectiveness and side effects on model capabilities.

\paragraph{Effectiveness of Editing.}
The effectiveness of editing is typically evaluated from four key properties using artificial benchmarks and simplified evaluation settings:
\begin{enumerate*}[label=\roman*)]
    \item[\ding{182}] \textit{reliability}, success rate of editing; 
    \item[\ding{183}] \textit{generalization}, adaptability of edited knowledge to paraphrased prompts;
    \item[\ding{184}] \textit{locality}, impact on irrelevant knowledge;
    \item[\ding{185}] \textit{portability}, applicability of edited knowledge in factual reasoning.
\end{enumerate*}
For detailed information, We refer readers to \citet{yao-etal-2023-editing}.
In addition to these basic metrics, domain-specific editing tasks have been introduced, e.g., privacy preservation \cite{wu-etal-2023-depn}, bias mitigation \cite{chen2024bias}, and harm injection \cite{chen2024harm}.

\paragraph{Side Effects of Editing.}
Recent research has also examined the potential side effects of editing on LLMs \cite{hoelscher-obermaier-etal-2023-detecting, li2024unveiling}.
While locality shares similar objectives, its limited evaluation scope fails to capture the full extent of editing side effects.
Recent studies \citep{yang-etal-2024-butterfly, gu-etal-2024-model, gupta-etal-2024-model} have revealed that model editing can significantly compromise LLMs' downstream tasks capabilities, motivating a growing research to mitigate such side effects \cite{corr24prune, fang2024alphaedit}.

\paragraph{Discussion.}
In contrast to prior efforts that either benchmark editing algorithms on synthetic datasets or analyze their side effects, this work offers the first systematic re-examination of model editing under realistic deployment conditions.
While AKEW \cite{wu-etal-2024-akew} shares our motivation of advancing model editing toward more realistic use cases, it pursues this goal by applying editing to a more complex task: unstructured editing.
Our study instead re-evaluates the effectiveness of existing editing methods on the same basic QA tasks adopted in prior work, but under a more rigorous and realistic evaluation protocol, revealing their limited practical utility and uncovering the pitfalls of traditional editing evaluation.

%% file: Section/3_Benchmark.tex
\section{QAEdit}  
\label{sec:qaedit}

\paragraph{Motivation.}
While existing work reports remarkable success of model editing techniques \cite{meng2023locating, wang2024wise}, their effectiveness in real-world applications remains unclear.
To rigorously examine their practical utility, we focus on the most fundamental and widely studied task of QA rather than more complex settings such as multi-hop and unstructured editing.
This choice is motivated by a simple premise: if current editing methods struggle on basic QA tasks, then they are unlikely to succeed in more challenging scenarios, whereas failure in such tasks does not entail failure on the basic QA task. %

Specifically, we apply editing methods to correct LLMs' errors in QA tasks and assess the improvement by re-evaluating edited LLMs on a standard QA evaluation framework, lm-evaluation-harness \cite{eval-harness}.

\paragraph{Benchmark Preparation.}
Since existing editing benchmarks are not derived from or aligned with mainstream QA tasks, we introduce QAEdit, a tailored benchmark to rigorously assess model editing in real-world QA. 
Specifically, QAEdit is constructed from three widely-used QA datasets with broad real-world coverage:  Natural Questions \cite{kwiatkowski2019nq}, TriviaQA \cite{joshi-etal-2017-triviaqa}, and SimpleQA \cite{wei2024measuringshortformfactualitylarge}.
Details about these datasets are provided in Appendix~\ref{apd:qa_data_intro}.

\input{Fig/QAEdit_Example}

\input{Fig/eval_framework_tikz}
\input{Table/pre_investigation}

While these benchmarks provide questions and answers as \textit{edit prompts} and \textit{targets} respectively, they lack essential fields that mainstream editing methods require for editing and evaluation.
To obtain required \textit{subjects} for editing, we employ GPT-4 (gpt-4-1106-preview) to extract them directly from the questions.
To align with the previous editing evaluation protocol, we evaluate:
\begin{enumerate*}[label=\roman*)]
\item \textit{reliability} using original edit prompts; 
\item \textit{generalization} through GPT-4 paraphrased prompts; and 
\item \textit{locality} using unrelated QA pairs from ZsRE locality set\footnote{We exclude \textit{portability} evaluation as it concerns reasoning rather than our focus on knowledge updating in real-world.}.
\end{enumerate*}

As a result, QAEdit contains 19,249 samples across ten categories, ensuring diverse coverage of QA scenarios. 
Figure~\ref{fig:QAedit_example} shows a QAEdit entry with all fields.
Dataset construction and dataset statistics are detailed in Appendix~\ref{apd:benchmark}.

\paragraph{Preliminary Study.}
We conduct single-edit experiments on Llama-2-7b-chat's failed questions in QAEdit (detailed in \S\ref{sec:single_edit}). 
As shown in Table~\ref{tab:pre_invest}, after applying SOTA editing methods, the edited models achieve only 38.5\% average accuracy under QA evaluation, far below previously reported results \cite{meng2023massediting, wang2024wise}.
This raises a critical question: \textit{Is the performance degradation attributed to the real-world complexity of QAEdit, or to real-world QA evaluation?}

%% file: Fig/QAEdit_Example.tex
\begin{figure}[t]
\centering
    \begin{tcolorbox}[
    right=0pt, left=0pt, top=0pt, bottom=0pt,
    toptitle=0.2mm, bottomtitle=0.2mm,
    colback=white,
    coltitle=white,
    colbacktitle=matisse,
    colframe=matisse,
    title=, %
    center title]
    \fontsize{6.5pt}{8pt}\selectfont
    \begin{minted}[autogobble,numberblanklines=false,breaklines]{json}
 "Edit Prompt"      : "To whom was Grete Stern married?",
 "Edit Target"      : "Horacio Coppola",
 "Subject"          : "Grete Stern",
 "Rephrased Prompt" : "Who was the spouse of Grete Stern?",
 "Locality Prompt"  : "When was the clock tower built in London?",
 "Locality Answer"  : "1859"
    \end{minted}
    \end{tcolorbox}
    \captionsetup{skip=4pt}
    \caption{An example from QAEdit.}
    \label{fig:QAedit_example}
\end{figure}

%% file: Fig/eval_framework_tikz.tex
\tikzset{
FARROW/.style={arrows={-{Latex[length=0.8mm, width=0.64mm]}}, rounded corners=0.6},
d_FARROW/.style={arrows={{Latex[length=0.8mm, width=0.64mm]}-{Latex[length=0.8mm, width=0.64mm]}}},
arrow1/.style={arrows={-{Latex[length=0.5mm, width=.4mm]}}, line width=0.4pt},
DFARROW/.style={arrows={{Latex[length=1.25mm, width=1.mm]}-{Latex[length=1.25mm, width=1.mm]}}},
hid_state/.style = {circle, fill=puerto_rico, minimum width=0.7em, align=center, inner sep=0, outer sep=0, font=\scriptsize},
pos_emb/.style = {circle, fill=flamingo!72, minimum width=1.8em, align=center, inner sep=0, outer sep=0, font=\scriptsize},
mha/.style = {rectangle, fill=sunset_orange, minimum width=0.7em, minimum height=0.7em, align=center, inner sep=0, outer sep=0, font=\scriptsize},
llm/.style = {rectangle, fill=none, draw, minimum width=11.5em, minimum height=6em, align=center, rounded corners=3}, %
mlp/.style = {diamond,
    inner sep=0, outer sep=0,
    minimum width=0.8em,
    minimum height=0.8em, fill=sushi, align=center},
project/.style = {rectangle, fill=hint_green, minimum width=3.6em, minimum height=1.1em, inner sep=1pt, outer sep=1pt, align=center, rounded corners=1.5, font=\small},
text_node/.style = { inner sep=2pt, outer sep=0pt, font=\fontsize{7.2pt}{9pt}\selectfont},
}

\tikzset{
  pics/self_att/.style={
      code={
          \node[hid_state] (-hs) at (0,0)  {};
          \node[mha, above=0.1em of -hs, xshift=0.7em] (-mha) {};
          \node[mlp, above=0.5em of -mha] (-mlp) {};

          \draw [FARROW, puerto_rico, thick] ($(-hs.north)+(0, .11em)$) to ($(-hs.north)+(0, 2.6em)$);
          \draw [arrow1, sunset_orange] ($(-mha.north)+(0, .01em)$) |- ++(-0.65em, 0.16em);
          \draw [arrow1, sushi] ($(-mha.north)+(-0.65em, 0.28em)$) -| ($(-mlp.south)+(0, -.01em)$);
          \draw [arrow1, sushi] ($(-mlp.north)+(0, .02em)$) |- ++(-0.65em, 0.14em);
          \draw [arrow1, sunset_orange] ($(-hs.east)+(0.01em, 0)$) -| ($(-mha.south)+(0, -.01em)$);
          \draw[sunset_orange, line width=0.5pt] (-hs) + (180:0.42em) arc (180:0:0.42em);
        }
    }
}

\tikzset{
  pics/my_dot/.style args={[#1]}{
    code={
      \tikzset{dot_color/.style={draw=#1}}
      \node[circle, minimum size=0.1em, align=center, inner sep=0, outer sep=0, draw, dot_color] (o1) at (0, 0) {};
      \node[circle, minimum size=0.1em, align=center, inner sep=0, outer sep=0, draw, dot_color] (o2) at ($(o1.north) + (0, .2em)$) {};
      \node[circle, minimum size=0.1em, align=center, inner sep=0, outer sep=0, draw, dot_color] (o3) at ($(o1.north) + (0, .4em)$) {};
    }
  },
  pics/my_dot/.default=[puerto_rico]
}

\tikzset{
  pics/my_cdot/.style args={[#1]}{
    code={
      \tikzset{dot_color/.style={fill=#1}}
      \node[circle, minimum size=0.12em, inner sep=0, outer sep=0, dot_color] (o1) at (0, 0) {};
      \node[circle, minimum size=0.12em, inner sep=0, outer sep=0, dot_color] at ($(o1.center) + (.2em, 0)$) {};
      \node[circle, minimum size=0.12em, inner sep=0, outer sep=0, dot_color] at ($(o1.center) + (.4em, 0)$) {};
    }
  },
  pics/my_cdot/.default=[sunset_orange]
}

\begin{figure*}
\begin{subfigure}[b]{0.48\textwidth}
  \begin{tikzpicture}

    \begin{scope}[opacity=0.15]
      \pic[name=sa10, local bounding box=sa10] at (0, 0) {self_att};
      \pic[name=sa11, local bounding box=sa11] at ($(sa10-hs) + (2.2em,0)$) {self_att};
      \pic[name=sa12, local bounding box=sa12] at ($(sa11-hs) + (2.8em,0)$) {self_att};
      
    \end{scope}

    \node[text_node,
      draw,
      rounded corners=1.5,
      anchor=north west,
      text width=6em,
    ]
    (ip2) at ($(sa10-hs) + (-0.6em, -1.3em)$) {\hspace*{0.03em} Who wrote the song \hspace*{0.1em}~``\textit{If I Were a Boy}'' ?};
    \node[text_node]
    (input_label) at ($(ip2.north) + (0, -2.4em)$) {\textcolor{flamingo}{\ding{202}} \bf context-free input};

    \node[text_node,
      anchor=north west]
    (de0) at ($(ip2.north east) + (1em,0)$) {\texttt{<BOS>}};
    \node[text_node,
      anchor=north west]
    (de1) at ($(de0.north east) + (0.15em,0)$) {BC};
    \node[text_node,
      anchor=north west]
    (de2) at ($(de1.north east) + (0.15em,0)$) {Jean};
    \node[text_node,
      anchor=north west]
    (de3) at ($(de2.north east) + (0.15em,0)$) {and};
    \node[text_node,
      anchor=north west]
    (de4) at ($(de3.north east) + (0.15em,0)$) {Toby};
    
    \draw [semithick, decorate, decoration={brace, amplitude=5pt, mirror}] ([yshift=-0.6em, xshift=-0.4em] de1.south west) -- ([yshift=-0.6em, xshift=6.2em] de1.south west) node[text_node, midway,yshift=-0.95em] (tf) {\textcolor{flamingo}{\ding{203}} \bf teacher forcing};
    
    \foreach \x in {0,1,2}
    {
    \draw [FARROW, puerto_rico, thick] ($(sa1\x-hs.south)+(0, -0.92em)$) to ++(0, 0.9em);
    }

    \begin{scope}[opacity=0.3]
      \foreach \x/\y in {0/5, 1/6, 2/7, 3/8, 4/9}
        {
          \pic[name=sa1\y, local bounding box=sa1\y] at ($(de\x |- sa10-hs)$){self_att};
          \pic[name=vdot\y, local bounding box=vdot\y] at ($(sa1\y-hs.north) + (0, 2.7em)$) {my_dot};
        }

      \foreach \x/\y in {5/6, 6/7, 7/8, 8/9}
        {
          \draw [arrow1, sunset_orange] ($(sa1\x-hs.east)+(0.04em, 0)$) -- ($(sa1\y-hs.west)+(-0.08em, 0)$);
        }
       
        \foreach \x in {5,6,7,8,9}
      {
        \node[hid_state] (sa3\x-hs) at ($(sa1\x-hs) + (0, 4.4em)$) {};
        \draw [FARROW, puerto_rico, thick] ($(vdot\x.north)+(0, 0.06em)$) to ($(sa3\x-hs.south)+(0, -0.04em)$);
      }
       
    \end{scope}
    
    \foreach \x in {5,...,9}
    {
    \draw [FARROW, puerto_rico, thick] ($(sa1\x-hs.south)+(0, -1.1em)$) to ++(0, 1.08em);
    }

    \begin{scope}[opacity=0.15]
      
      \foreach \x in {0,1,2}
      {
        \node[hid_state] (sa3\x-hs) at ($(sa1\x-hs) + (0, 4.4em)$) {};
      }

      \foreach \x in {1}
        {
          \draw [arrow1, sunset_orange] ($(sa\x0-hs.east)+(0.04em, 0)$) -- ($(sa\x1-hs.west)+(-0.08em, 0)$);
          \pic[name=cdot\x, local bounding box=cdot\x] at ($(sa\x1-hs.east)!0.52!(sa\x2-hs.west)$) {my_cdot=[puerto_rico]};
          \draw [arrow1, sunset_orange] ($(sa\x1-hs.east)+(0.04em, 0)$) -- ($(cdot\x.west)+(-0.04em, 0) $); %
          \draw [arrow1, sunset_orange] ($(cdot\x.east)+(0.08em, 0)$) -- ($(sa\x2-hs.west) +(-0.08em, 0) $); %
        }

      \pic[name=cdot1-1, local bounding box=cdot1-1] at ($(cdot1o1 |- sa11-mha)$) {my_cdot};
      \pic[name=cdot1-2, local bounding box=cdot1-2] at ($(cdot1o1 |- sa11-mlp)$) {my_cdot=[sushi]};

      \foreach \x in {0,1,2}
        {
          \pic[name=vdot\x, local bounding box=vdot\x] at ($(sa1\x-hs.north) + (0, 2.7em)$) {my_dot};
          \draw [FARROW, puerto_rico, thick] ($(vdot\x.north)+(0, 0.06em)$) to ($(sa3\x-hs.south)+(0, -0.04em)$);
        }
        
     \pic[name=cdot_io_0, local bounding box=cdot_io_0] at ($(sa12-hs.east)!0.5!(sa15-hs.west)$) {my_cdot=[puerto_rico]};
    \foreach \x/\y in {1/sa12-mha}
      {
        \pic[name=cdot_io_\x, local bounding box=cdot_io_\x] at ($(cdot_io_0o1 |- \y)$) {my_cdot};
      }

     \foreach \x/\y in {2/sa12-mlp}
      {
        \pic[name=cdot_io_\x, local bounding box=cdot_io_\x] at ($(cdot_io_0o1 |- \y)$) {my_cdot=[sushi]};
      }
      
      \pic[name=cdot_io_30, local bounding box=cdot_io_30] at ($(cdot1o1 |- sa30-hs)$) {my_cdot=[puerto_rico]};
      
      \pic[name=cdot_io_31, local bounding box=cdot_io_31] at ($(cdot_io_0o1 |- sa30-hs)$) {my_cdot=[puerto_rico]};
    
    \draw [arrow1, sunset_orange] ($(sa12-hs.east)+(0.04em, 0)$) -- ($(cdot_io_0.west)+(-0.04em, 0) $);

    \end{scope}

	\begin{scope}[opacity=0.3]
     \draw [arrow1, sunset_orange] ($(cdot_io_0.east)+(0.08em, 0)$) -- ($(sa15-hs.west) +(-0.08em, 0) $);

	\end{scope}
    
    \node[text_node,
      anchor=base,
      text depth=.15em,
      xshift=-0.3em,
      ]
    (de_out1) at ($(sa15-hs.north) + (0, 5.5em)$) {Beyonc\'e};
    
    \node[text_node,
      anchor=base, 
      text depth=.15em,
      ]
    (de_out2) at ($(sa16-hs.north |- de_out1.base)$) {Jean};
    
    \node[text_node,
  anchor=base, 
      text depth=.15em,
  ]
(de_out3) at ($(sa17-hs.north |- de_out1.base)$) {is};

    \node[text_node,
  anchor=base, 
      text depth=.15em,
  ]
(de_out4) at ($(sa18-hs.north |- de_out1.base)$) {Toby};
    \node[text_node,
  anchor=base, 
      text depth=.15em,
  ]
(de_out5) at ($(sa19-hs.north |- de_out1.base)$) {Gad};

    \node[text_node,
       anchor=base, 
      text depth=.15em,
      ]
    (gt5) at ($(sa15-hs.north) + (0, 7.7em)$) {BC};

    \foreach \x/\y in {6/Jean, 7/and, 8/Toby, 9/Gad}
    {
    \node[text_node,
      anchor=base, 
      text depth=.15em,
      ]
    (gt\x) at ($(sa1\x-hs.south |- gt5.base)$) {\y};
    }
    
    \node[text_node,
      anchor=base east,
      align=right,
      text depth=.15em,
      text width=2em,
      color=tuatara,
      font=\fontsize{4pt}{5pt}\selectfont \linespread{0.4}\selectfont
      ]
    (gt) at ($(de_out1.west) + (0, 2.65em) $) {\bf Ground Truth:};
    
    \node[text_node,
      anchor=base east,
      text depth=.15em,
      color=tuatara,
      font=\fontsize{4pt}{5pt}\selectfont
      ]
    (op_label) at ($(de_out1.west) + (0.05em, 0.15em) $) {\bf Output:};

	\begin{scope}[on background layer]
	    \node[%
	          draw=tuatara!42, thin,
	          rounded corners=1pt,
	          fit={(gt5) (gt9) ($(gt5.west) + (-1em,0.02em)$)  ($(gt9.east) + (0.3em, 0.55em)$)},
	          inner sep=-0.5pt,
	    ] (gt_box) {};
	    \node[%
	          draw=tuatara!42, thin,
	          rounded corners=1pt,
	          fit={(de_out1) (de_out5) ($(de_out5.east) + (.3em,0.55em)$)},
	          inner sep=-0.5pt,
	    ] (output_box) {};
	\end{scope}

	 \foreach \x/\y in {5, 6, 7, 8, 9}
      {
         \draw [FARROW, puerto_rico, thick] ($(sa3\x-hs.north)+(0, 0.04em)$) to ++(0, 1.05em) ;
      }

      \node[regular polygon, regular polygon sides=8, draw=flamingo!200, fill=flamingo, text=white, minimum size=0.2em, inner sep=0pt, outer sep=0pt] (stop) at ($(gt9.east) + (1.3em, -10em) $) {\fontsize{4pt}{5pt}\selectfont\bf STOP};
      
      \draw [FARROW, flamingo, thick, dash pattern=on 1pt off 0.8pt] (gt_box.east) to ($(gt_box.east -| stop.north)$) to node[midway, sloped=true, color=black, font=\scriptsize, yshift=0.5em, xshift=0.1em] (stop_label) {\bf ground truth length} (stop.north) ;
      \node[] at ($(stop_label.west) + (0, 0.06em)$) {\small \textcolor{flamingo}{\ding{204}}};

      \foreach \x/\y/\z in {5/\ding{56}/flamingo, 6/\ding{52}/shakespeare, 7/\ding{56}/flamingo, 8/\ding{52}/shakespeare, 9/\ding{52}/shakespeare}
      {
         \draw [d_FARROW, dotted, dash pattern=on 0.6pt off 0.48pt] ($(gt\x.south)+(0, 0.32em)$) -- node[midway, sloped=false, color=\z, font=\scriptsize] (m\x) {\y}  ++(0, -1.82em) ;
      }
     
      \node[densely dotted,
          draw=flamingo,
          fit={(m5) (m9) },
          inner sep=-2.5pt,
	    ] (m_box) {};
	 
	 \draw [FARROW, red, densely dotted] ($(m_box.west)+(-0.1em, 0)$) to ++(-0.6em, 0);
     \draw [FARROW, red, densely dotted] ($(m_box.west)+(-2.5em, 0)$) to ++(0.9em, 0);
     
     \node[text_node,
      anchor=center,
      align=center,
      text=flamingo!62!red,
    ]
    (result) at ($(gt9.east) + (-8.8em, -1.1em)$) {3/5};

     \draw [FARROW] ($(gt_box.west)+(0, 0)$) to ++(-3em, 0) to ++(0, -0.55em);
     \draw [FARROW] ($(output_box.west)+(0, 0)$) to ++(-3em, 0) to ++(0, 0.55em);
     \node[project,
      anchor=center,
      align=center,
      font=\fontsize{7.2pt}{9pt}\selectfont
    ]
    (metric) at ($(gt.west) + (-0.7em, -1.6em)$) {match ratio};
    \node[] at ($(metric.west) + (-0.2em, 0em)$) {\fontsize{7.2pt}{9pt}\selectfont \textcolor{flamingo}{\ding{205}}};

    \begin{scope}[on background layer]
    
    \coordinate (target1) at ($(gt5.east)+(0, -10.9em)$);

    \draw [FARROW, red] ($(gt5.east)+(-0.2em, 0)$) to ($(gt5.east)+(0.4em, 0)$) to ($(gt5.east)+(0.4em, -10.9em)$) to ($(target1 -| de1.south)$) to ($(de1.south)+(0, 0.2em)$) ;
    \draw [FARROW, red
    ] ($(gt6.east)+(-0.2em, 0)$) to ++ (0.3em, 0) to ++(0, -10.9em) to ($(target1 -| de2.south)$) to ($(de2.south)+(0, 0.2em)$);
    \draw [FARROW, red
    ] ($(gt7.east)+(-0.25em, 0)$) to ++ (0.45em, 0) to ++(0, -10.9em) to ($(target1 -| de3.south)$) to ($(de3.south)+(0, 0.2em)$);
    
    \draw [FARROW, red
    ] ($(gt8.east)+(-0.2em, 0)$) to ++ (0.3em, 0) to ++(0, -10.9em) to ($(target1 -| de4.south)$) to ($(de4.south)+(0, 0.35em)$);
    	
    \end{scope}

    \begin{scope}[on background layer]
    \node[fill=gallery!42,
          rounded corners=2pt,
          fit={(sa10)(sa15)(sa16)(sa17)(sa18)(sa19)(sa30-hs)},
          inner sep=3pt,
          ] (dashedBox) {};
          
     \draw [dashed, flamingo] ($(cdot_io_0.north)+(-0.4em, -3.8em)$) to ($(cdot_io_31.north)+(-0.4em, 0.9em)$) ;
     
     \end{scope}
    
    \node[
          draw,
          rounded corners=2pt,
          fit={(sa10)(sa15)(sa16)(sa17)(sa18)(sa19)(sa30-hs)},
          inner sep=3pt,
          ] {};
     \node[
          rounded corners=2pt, opacity=0.8,
          fit={(sa10)(sa15)(sa16)(sa17)(sa18)(sa19)(sa30-hs)},
          inner sep=7pt,
          align=center, text height=24pt, text depth=0.9em,
          ]  {\fontsize{24}{30}\selectfont Edited $\,$ LLM};

  \end{tikzpicture} %
  \caption{synthetic evaluation framework}
  \label{fig:van_eval}
\end{subfigure}
\begin{subfigure}[b]{0.48\textwidth}
  \begin{tikzpicture}

    \begin{scope}[opacity=0.15]
      \pic[name=sa10, local bounding box=sa10] at (0, 0) {self_att};
      \pic[name=sa11, local bounding box=sa11] at ($(sa10-hs) + (2.2em,0)$) {self_att};
      \pic[name=sa12, local bounding box=sa12] at ($(sa11-hs) + (3em,0)$) {self_att};
    \end{scope}

    \node[text_node,
      draw,
      rounded corners=1.5,
      anchor=north west,
      text width=7.2em
    ]
    (ip2) at ($(sa10-hs) + (-0.6em, -1.3em)$) {\texttt{\{Context\}} Who wrote the song ``\textit{If I Were a Boy}'' ?};
    
    \begin{scope}[on background layer]
    	\node[project,
      rectangle, fill=flamingo!62, minimum width=3em, minimum height=0.78em, rounded corners=1,
    ]
    (c_bg) at ($(ip2.west) + (1.55em, 0.42em)$) {};
    \end{scope}

    \node[text_node]
    (input_label) at ($(ip2.north) + (0, -2.4em)$) {\textcolor{flamingo}{\ding{202}} \bf context-guided input};

    \node[text_node,
      anchor=north west]
    (de0) at ($(ip2.north east) + (1em,0)$) {\texttt{<BOS>}};
    \node[text_node,
      anchor=north west]
    (de1) at ($(de0.north east) + (0,0)$) {Beyonc\'e};
    \node[text_node,
      anchor=north west]
    (de2) at ($(de1.north east) + (0.15em,0)$) {is};
    \node[text_node,
      anchor=north west]
    (de3) at ($(de2.north east) + (0.8em,0)$) {the};
    \node[text_node,
      anchor=north west]
    (de4) at ($(de3.north east) + (0.35em,0)$) {writer};
    
\draw [semithick, decorate, decoration={brace, amplitude=5pt, mirror}] ([yshift=-0.6em, xshift=-1.6em] de1.south west) -- ([yshift=-0.6em, xshift=7.5em] de1.south west) node[text_node, midway,yshift=-0.8em] (tf) {\textcolor{flamingo}{\ding{203}} \bf autoregressive decoding};
    
    \foreach \x in {0,1,2}
    {
    \draw [FARROW, puerto_rico, thick] ($(sa1\x-hs.south)+(0, -0.92em)$) to ++(0, 0.9em);
    }

    \begin{scope}[opacity=0.3]
      \foreach \x/\y in {0/5, 1/6, 2/7, 3/8, 4/9}
        {
          \pic[name=sa1\y, local bounding box=sa1\y] at ($(de\x |- sa10-hs)$){self_att};
          \pic[name=vdot\y, local bounding box=vdot\y] at ($(sa1\y-hs.north) + (0, 2.7em)$) {my_dot};
        }

      \foreach \x/\y in {5/6, 6/7, 7/8, 8/9}
        {
          \draw [arrow1, sunset_orange] ($(sa1\x-hs.east)+(0.04em, 0)$) -- ($(sa1\y-hs.west)+(-0.08em, 0)$);
        }
        
        \foreach \x in {5,6,7,8,9}
      {
        \node[hid_state] (sa3\x-hs) at ($(sa1\x-hs) + (0, 4.4em)$) {};
        \draw [FARROW, puerto_rico, thick] ($(vdot\x.north)+(0, 0.06em)$) to ($(sa3\x-hs.south)+(0, -0.04em)$);
      }
       
    \end{scope}
    
    \foreach \x in {5,...,9}
    {
    \draw [FARROW, puerto_rico, thick] ($(sa1\x-hs.south)+(0, -1.1em)$) to ++(0, 1.08em);
    }

    \begin{scope}[opacity=0.15]

      \foreach \x in {0,1,2}
      {
        \node[hid_state] (sa3\x-hs) at ($(sa1\x-hs) + (0, 4.4em)$) {};
        
      }

      \foreach \x in {1}
        {
          \draw [arrow1, sunset_orange] ($(sa\x0-hs.east)+(0.04em, 0)$) -- ($(sa\x1-hs.west)+(-0.08em, 0)$);
          \pic[name=cdot\x, local bounding box=cdot\x] at ($(sa\x1-hs.east)!0.52!(sa\x2-hs.west)$) {my_cdot};

          \draw [arrow1, sunset_orange] ($(sa\x1-hs.east)+(0.04em, 0)$) -- ($(cdot\x.west)+(-0.04em, 0) $); %
          \draw [arrow1, sunset_orange] ($(cdot\x.east)+(0.08em, 0)$) -- ($(sa\x2-hs.west) +(-0.08em, 0) $); %
        }

      \pic[name=cdot1-1, local bounding box=cdot1-1] at ($(cdot1o1 |- sa11-mha)$) {my_cdot};
      \pic[name=cdot1-2, local bounding box=cdot1-2] at ($(cdot1o1 |- sa11-mlp)$) {my_cdot=[sushi]};

      \foreach \x in {0,1,2}
        {
          \pic[name=vdot\x, local bounding box=vdot\x] at ($(sa1\x-hs.north) + (0, 2.7em)$) {my_dot};
          \draw [FARROW, puerto_rico, thick] ($(vdot\x.north)+(0, 0.06em)$) to ($(sa3\x-hs.south)+(0, -0.04em)$);
        }

    \pic[name=cdot_io_0, local bounding box=cdot_io_0] at ($(sa12-hs.east)!0.5!(sa15-hs.west)$) {my_cdot=[puerto_rico]};
    \foreach \x/\y in {1/sa12-mha}
      {
        \pic[name=cdot_io_\x, local bounding box=cdot_io_\x] at ($(cdot_io_0o1 |- \y)$) {my_cdot};
      }

     \foreach \x/\y in {2/sa12-mlp}
      {
        \pic[name=cdot_io_\x, local bounding box=cdot_io_\x] at ($(cdot_io_0o1 |- \y)$) {my_cdot=[sushi]};
      }
      
      \pic[name=cdot_io_30, local bounding box=cdot_io_30] at ($(cdot1o1 |- sa30-hs)$) {my_cdot=[puerto_rico]};
      
      \pic[name=cdot_io_31, local bounding box=cdot_io_31] at ($(cdot_io_0o1 |- sa30-hs)$) {my_cdot=[puerto_rico]};
    
    \draw [arrow1, sunset_orange] ($(sa12-hs.east)+(0.04em, 0)$) -- ($(cdot_io_0.west)+(-0.04em, 0) $);
    \end{scope}

	\begin{scope}[opacity=0.3]
     \draw [arrow1, sunset_orange] ($(cdot_io_0.east)+(0.08em, 0)$) -- ($(sa15-hs.west) +(-0.08em, 0) $);
	\end{scope}

    \node[text_node,
      anchor=base,        %
      text depth=.15em,
      xshift=-0.3em,
      ]
    (de_out1) at ($(sa15-hs.north) + (0, 5.5em)$) {Beyonc\'e};
    
    \node[text_node,
      anchor=base,        %
      text depth=.15em,
      ]
    (de_out2) at ($(sa16-hs.north |- de_out1.base)$) {is};
    
    \node[text_node,
  anchor=base,        %
      text depth=.15em,
  ]
(de_out3) at ($(sa17-hs.north |- de_out1.base)$) {the};

    \node[text_node,
  anchor=base,        %
      text depth=.15em,
  ]
(de_out4) at ($(sa18-hs.north |- de_out1.base)$) {writer};
    \node[text_node,
      anchor=base,        %
      text depth=.15em,
      font=\fontsize{4.4pt}{5pt}\selectfont,
      xshift=0.4em,
      opacity=0.5,
  ]
(de_out5) at ($(sa19-hs.north |- de_out1.base)$) {<|endoftext|>};
\begin{scope}[on background layer]
    	\node[project,
      rectangle, fill=flamingo!60, minimum width=2.2em, minimum height=0.6em, rounded corners=1,
    ]
    (o_bg) at ($(de_out5) + (0, 0.05em)$) {};
    \end{scope}

    \node[text_node,
       anchor=base,
       text opacity=0,
      text depth=.15em,
      ]
    (gt5) at ($(sa15-hs.north) + (0, 7.7em)$) {BC};
     \node[text_node,
       anchor=base,
      text depth=.15em,
      ]
    (gt0) at ($(sa15-hs.north) + (1.6em, 7.7em)$) {BC Jean and Toby Gad};

    \foreach \x/\y in {6/Jean, 7/and, 8/Toby, 9/Gad}
    {
    \node[text_node,
       text opacity=0,
      anchor=base,
      text depth=.15em,
      ]
    (gt\x) at ($(sa1\x-hs.south |- gt5.base)$) {\y};
    }
    
    \node[text_node,
      anchor=base east,
      align=right,
      text depth=.15em,
      text width=2em,
      color=tuatara,
      font=\fontsize{4pt}{5pt}\selectfont \linespread{0.4}\selectfont
      ]
    (gt) at ($(de_out1.west) + (-0.2em, 2.65em) $) {\bf Ground Truth:};
    
    \node[text_node,
      anchor=base east,
      text depth=.15em,
      color=tuatara,
      font=\fontsize{4pt}{5pt}\selectfont
      ]
    (op_label) at ($(de_out1.west) + (-0.15em, 0.15em) $) {\bf Output:};

	\begin{scope}[on background layer]
	    \node[fill=athens_gray!42, 
	          draw=tuatara!42, thin,
	          rounded corners=1pt,
	          fit={(gt0)  ($(gt0.west) + (-0.2em, 0)$)  ($(gt0.east) + (0.2em, 0.55em)$)},
	          inner sep=-0.5pt,
	    ] (gt_box) {};
	    \node[fill=athens_gray!42, 
	          draw=tuatara!42, thin,
	          rounded corners=1pt,
	          fit={(de_out1) (de_out4) ($(de_out1.west) + (-0.2em,0)$) ($(de_out4.east) + (0.2em,0)$)},
	          inner sep=-0.5pt,
	    ] (output_box) {};
	\end{scope}

	 \foreach \x/\y in {5, 6, 7, 8, 9}
      {
         \draw [FARROW, puerto_rico, thick] ($(sa3\x-hs.north)+(0, 0.04em)$) to ++(0, 1.05em) ;
      }

     \node[regular polygon, regular polygon sides=8, draw=flamingo!200, fill=flamingo, text=white, minimum size=0.2em, inner sep=0pt, outer sep=0pt] (stop) at ($(gt9.east) + (1.25em, -10em) $) {\fontsize{4pt}{5pt}\selectfont\bf STOP};
      
     \coordinate (stp_p) at ($(de_out5.east)+ (0, 0.05em) $);
      
      \draw [FARROW, flamingo, thick, dash pattern=on 1pt off 0.8pt] ($(de_out5.east)+ (-0.1em, 0.05em) $) to ($(stp_p -| stop.north)$) to node[midway, sloped=true, color=black, font=\scriptsize, yshift=0.4em, xshift=-0.2em] (stop_label) {\bf  natural stopping criteria} (stop.north) ;
      \node[text_node] at ($(stop_label.west) + (0, 0)$) { \textcolor{flamingo}{\ding{204}}};

     \node[
      anchor=center,
      align=center,
    ]
    (metric) at ($(gt.west) + (-0.6em, -1.6em)$) {\textcolor{ship_gray}{\large \faIcon{robot}}};
    \node[text width=2.3em, text depth=.1em]
    (metric_label) at ($(metric.west) + (-0.7em, 0.3em)$) {\fontsize{6pt}{1pt}\selectfont  \bfseries LLM-as-};
    \node[text width=2.3em, text depth=.1em, align=left, anchor=west]
    (metric_label1) at ($(metric_label.west) + (0, -0.6em)$) {\fontsize{6pt}{1pt}\selectfont  \bfseries  a-Judge};
    \node[text_node]
    (metric_an) at ($(metric_label.west) + (0, -0.3em)$) {\textcolor{flamingo}{\ding{205}}};
    \node[text_node,
     text=flamingo!62!red,
    ]
    (result) at ($(metric.east) + (1em, 0)$) {0};

     \draw [FARROW, red, densely dotted] ($(result.west)+(-0.9em, 0)$) to ++(1.1em, 0);
    \draw [FARROW] ($(gt_box.west)+(0, 0)$) to ++(-3em, 0) to ++(0, -0.55em);
     \draw [FARROW] ($(output_box.west)+(0, 0)$) to ++(-3em, 0) to ++(0, 0.55em);

    \begin{scope}[on background layer]
    
    \coordinate (target1) at ($(de_out1.east)+(0, -8.7em)$);

    \draw [FARROW, red] ($(de_out1.east)+(-0.2em, 0)$) to ($(de_out1.east)+(0, 0)$) to ++(0, -8.7em) to ($(target1 -| de1.south)$) to ($(de1.base)+(0, -0.1em)$) ;
    \draw [FARROW, red] ($(de_out2.east)+(-0.2em, 0)$) to ++ (1.1em, 0) to ++(0, -8.7em) to ($(target1 -| de2.south)$) to ($(de2.base)+(0, -0.1em)$);
    \draw [FARROW, red] ($(de_out3.east)+(-0.2em, 0)$) to ++ (0.35em, 0) to ++(0, -8.7em) to ($(target1 -| de3.south)$) to ($(de3.base)+(0, -0.1em)$);  
    \draw [FARROW, red] ($(de_out4.east)+(-0.2em, 0)$) to ++ (0.2em, 0) to ++(0, -8.7em) to ($(target1 -| de4.south)$) to ($(de4.base)+(0, -0.1em)$);
    	
    \end{scope}

    \begin{scope}[on background layer]
    \node[fill=gallery!42,
          rounded corners=2pt,
          fit={(sa10)(sa15)(sa16)(sa17)(sa18)(sa19)(sa30-hs)},
          inner sep=3pt,
          ] (dashedBox) {};
          
     \draw [dashed, flamingo] ($(cdot_io_0.north)+(0, -3.8em)$) to ($(cdot_io_31.north)+(0, 1.2em)$) ;
     
     \end{scope}
    
    \node[
          draw,
          rounded corners=2pt,
          fit={(sa10)(sa15)(sa16)(sa17)(sa18)(sa19)(sa30-hs)},
          inner sep=3pt,
          ] {};
     \node[
          rounded corners=2pt, opacity=0.8,
          fit={(sa10)(sa15)(sa16)(sa17)(sa18)(sa19)(sa30-hs)},
          inner sep=7pt,
          align=center, text height=24pt, text depth=0.9em,
          ]  {\fontsize{24}{30}\selectfont Edited $\,$ LLM};

  \end{tikzpicture} %
  \caption{\textsc{Wild} evaluation framework}
  \label{fig:prac_eval}
\end{subfigure}
\caption{Illustration of synthetic and \textsc{Wild} evaluation frameworks for measuring reliability, generalization, and locality. Each framework comprises four key modules:\textcolor{flamingo}{\ding{202}}$\,$\textit{input}, \textcolor{flamingo}{\ding{203}}$\,$\textit{generation strategy}, \textcolor{flamingo}{\ding{204}}$\,$\textit{output truncation}, and \textcolor{flamingo}{\ding{205}}$\,$\textit{metric}. Here, we use LLM-as-a-Judge as an example metric to illustrate \textsc{Wild}, which supports various metrics.}
\label{fig:eval_frame}
\end{figure*}

%% file: Table/pre_investigation.tex
\begin{table}[t]
\centering
\setlength{\tabcolsep}{3.5pt}
\begin{adjustbox}{max width=\linewidth} 
\begin{tabular}{lrrrrrrc}
\toprule
\textbf{Method} & FT-M & MEND & ROME & MEMIT & GRACE & WISE & Avg. \\
\midrule
\textbf{Accuracy} & \num{0.611} & \num{0.333} & \num{0.585} & \num{0.552} & \num{0.012} & \num{0.216} & \num{0.385} \\
\bottomrule 
\end{tabular}
\end{adjustbox}
\caption{Accuracy of edited Llama-2-7b-chat on questions it failed before editing in QAEdit.}
\label{tab:pre_invest}
\end{table}

%% file: Section/4_Evaluation.tex
\section{A Tale of Two Evaluation Frameworks}
\label{sec:eval}

To identify the cause of this performance gap and guide further investigation, we first delve into the experimental setup of both editing (synthetic) and QA task (\textsc{Wild}) evaluations.
We abstract them into four key modules: \textit{input}, \textit{generation strategy}, \textit{output truncation}, and \textit{metric}.
This modular paradigm enables systematic comparison between the two evaluation frameworks, as shown in Figure~\ref{fig:eval_frame}.

\paragraph{Synthetic.}
We formalize the evaluation pipeline commonly used in prior model editing works \cite{yao-etal-2023-editing, wang2024wise} as \textbf{synthetic} evaluation framework, which implements the four modules in an idealized and overly simplified way (Figure~\ref{fig:van_eval}):
\begin{enumerate*}[label=\roman*)]
    \item[\ding{182}] \textbf{input}: using only question without additional context;
    \item[\ding{183}] \textbf{generation strategy}: employing teacher forcing to feed ground truth tokens as input during decoding\footnote{The code snippets of mainstream editing evaluations with teacher forcing are presented in Appendix~\ref{apd:code4tf}.};
    \item[\ding{184}] \textbf{output truncation}: truncating output to match the length of target answer;
    \item[\ding{185}] \textbf{metric}: using token-level match ratio between the target and generated answer as accuracy.
\end{enumerate*}

\paragraph{\textsc{Wild}.} We propose the \textsc{Wild} (\textbf{W}ithout \textbf{I}nterven-tion, \textbf{L}ive \textbf{D}ecoding) evaluation framework based on the standard QA evaluation protocol \cite{eval-harness}, which implements the core modules in a more realistic manner (Figure~\ref{fig:prac_eval}):
\begin{enumerate*}%
    \item[\ding{182}] \textbf{input}: prefixing question with contexts like task instructions;
    \item[\ding{183}] \textbf{generation strategy}: adopting autoregressive decoding, where each output serves as input for subsequent generation;
    \item[\ding{184}] \textbf{output truncation}: using predefined stop tokens (e.g., ``.'', ``\texttt{\textbackslash{}n}'', and ``\texttt{<|endoftext|>}'') as signal to terminate generation;
    \item[\ding{185}] \textbf{metric}: \textsc{Wild} supports evaluation metrics, including BERTScore \cite{zhangbertscore} and exact match (EM). Given its popularity and alignment with human judgment, we adopt LLM-as-a-Judge\footnote{Detailed prompt is provided in Appendix~\ref{apd:judge_prompt}.} \cite{li2024llmasjudge} as the primary metric to illustrate the framework and conduct our study. Additional metric discussions are provided in \S~\ref{sec:analysis_metric}.
\end{enumerate*}

Here, we use the basic QA task to instantiate the \textsc{Wild} evaluation framework, as our study focuses on improving the realism of evaluation, rather than increasing task complexity.
Notably, our proposed framework is task-agnostic and can be easily applied to more complex scenarios, including multi-hop and unstructured editing.

\input{Table/comp_evals}

\input{Table/main_exp_result_color}

\paragraph{Discussion.}
Table~\ref{tab:comp_evals} details the key differences between these evaluation frameworks. 
Previous synthetic evaluation has two types of critical limitations compared to \textsc{Wild} evaluation:
\begin{enumerate*}%
    \item[\ding{182}] \textbf{oversimplification}: context-free input overlooks the complexity and variability of practical queries, and match ratio rewards partial matches of incorrect answers;
    \item[\ding{183}] \textbf{unreasonableness}: teacher forcing generation and corresponding truncation to the target length leak ground truth information that should remain inaccessible during testing.
\end{enumerate*}
These artificial settings result in a significant gap between research on editing and its practical applications.

%% file: Table/comp_evals.tex
\begin{table}[t]
\centering
\begin{adjustbox}{max width=\linewidth} 
\begin{tabular}{l l l}
    \toprule
    \textbf{Module} & \textbf{synthetic} & \textsc{\textbf{Wild}} \\ 
    \midrule
    \textbf{Input} &   context-free & context-guided \\ 
    \textbf{Gen. Strategy} & teacher forcing & autoregressive decoding \\ 
    \textbf{Output Trunc.} & ground truth length & natural stopping criteria \\ 
    \textbf{Metric} &   match ratio & LLM-as-a-Judge / EM \\ 
    \bottomrule
\end{tabular}
\end{adjustbox}
\caption{Key settings of synthetic and \textsc{Wild} evaluation across all four modules.}
\label{tab:comp_evals}
\end{table}

%% file: Table/main_exp_result_color.tex
\begin{table*}[t]
\begin{minipage}{0.94\textwidth}
    \centering
    \renewcommand{\arraystretch}{1.05}
    \begin{adjustbox}{max width=\textwidth}
    \begin{tabular}{l lcc cc cc cc cc cc}
    \toprule
    & \multirow{3.8}{*}{\textbf{Method}} & \multicolumn{4}{c}{\textbf{ZsRE}} & \multicolumn{4}{c}{\textbf{\textsc{CounterFact}}} & \multicolumn{4}{c}{\textbf{QAEdit}} \\
    \cmidrule(lr){3-6}\cmidrule(lr){7-10}\cmidrule(lr){11-14} 
    & & \multicolumn{2}{c}{Reliability} & \multicolumn{2}{c}{Generalization} & \multicolumn{2}{c}{Reliability} & \multicolumn{2}{c}{Generalization} & \multicolumn{2}{c}{Reliability} & \multicolumn{2}{c}{Generalization} \\
     \cmidrule(lr){3-4} \cmidrule(lr){5-6} \cmidrule(lr){7-8} \cmidrule(lr){9-10} \cmidrule(lr){11-12} \cmidrule(lr){13-14}
    & & syn. & \textsc{Wild}  & syn. & \textsc{Wild} & syn.  & \textsc{Wild}  & syn. & \textsc{Wild} & syn. & \textsc{Wild} & syn.  & \textsc{Wild}  \\
     \midrule
\multirow{6.5}{*}{\rotatebox{90}{Llama-2-7b-chat}} &  FT-M    & \numedit{1.000} & \numrwe{0.562}{1.000} & \numedit{0.950} & \numrwe{0.470}{0.950} & \numedit{1.000} & \numrwe{0.867}{1.000} & \numedit{0.503} & \numrwe{0.426}{0.503} & \numedit{1.000} & \numrwe{0.611}{1.000} & \numedit{0.966} & \numrwe{0.560}{0.966} \\
& MEND    & \numedit{0.967} & \numrwe{0.288}{0.967} & \numedit{0.949} & \numrwe{0.244}{0.949} & \numedit{0.997} & \numrwe{0.478}{0.997} & \numedit{0.425} & \numrwe{0.183}{0.425} & \numedit{0.942} & \numrwe{0.333}{0.942} & \numedit{0.900} & \numrwe{0.328}{0.900} \\
& ROME  & \numedit{0.964} & \numrwe{0.741}{0.964} & \numedit{0.811} & \numrwe{0.656}{0.811} & \numedit{0.996} & \numrwe{0.836}{0.996} & \numedit{0.452} & \numrwe{0.420}{0.452} & \numedit{0.955} & \numrwe{0.585}{0.955} & \numedit{0.744} & \numrwe{0.411}{0.744} \\
& MEMIT   & \numedit{0.950} & \numrwe{0.685}{0.950} & \numedit{0.858} & \numrwe{0.634}{0.858} & \numedit{0.997} & \numrwe{0.797}{0.997} & \numedit{0.513} & \numrwe{0.460}{0.513} & \numedit{0.929} & \numrwe{0.552}{0.929} & \numedit{0.791} & \numrwe{0.450}{0.791} \\
& GRACE   & \numedit{0.986} & \numrwe{0.033}{0.986} & \numedit{0.319} & \numrwe{0.029}{0.319} & \numedit{0.998} & \numrwe{0.013}{0.998} & \numedit{0.114} & \numrwe{0.008}{0.114} & \numedit{0.983} & \numrwe{0.012}{0.983} & \numedit{0.383} & \numrwe{0.087}{0.383} \\
& WISE    & \numedit{0.999} & \numrwe{0.139}{0.999} & \numedit{0.973} & \numrwe{0.081}{0.973} & \numedit{0.999} & \numrwe{0.521}{0.999} & \numedit{0.612} & \numrwe{0.104}{0.612} & \numedit{0.998} & \numrwe{0.216}{0.998} & \numedit{0.877} & \numrwe{0.122}{0.877} \\
     \midrule
 \multirow{6.5}{*}{\rotatebox{90}{Mistral-7b}}  &  FT-M    & \numedit{1.000} & \numrwe{0.441}{1.000} & \numedit{0.824} & \numrwe{0.358}{0.824} & \numedit{1.000} & \numrwe{0.733}{1.000} & \numedit{0.330} & \numrwe{0.220}{0.330} & \numedit{1.000} & \numrwe{0.562}{1.000} & \numedit{0.862} & \numrwe{0.503}{0.862} \\
& MEND    & \numedit{0.977} & \numrwe{0.719}{0.977} & \numedit{0.963} & \numrwe{0.657}{0.963} & \numedit{0.820} & \numrwe{0.431}{0.820} & \numedit{0.355} & \numrwe{0.149}{0.355} & \numedit{0.903} & \numrwe{0.544}{0.903} & \numedit{0.895} & \numrwe{0.516}{0.895} \\
& ROME  & \numedit{0.757} & \numrwe{0.608}{0.757} & \numedit{0.717} & \numrwe{0.573}{0.717} & \numedit{0.965} & \numrwe{0.866}{0.965} & \numedit{0.466} & \numrwe{0.488}{0.466} & \numedit{0.845} & \numrwe{0.555}{0.845} & \numedit{0.735} & \numrwe{0.435}{0.735} \\
& MEMIT   & \numedit{0.868} & \numrwe{0.707}{0.868} & \numedit{0.842} & \numrwe{0.670}{0.842} & \numedit{0.962} & \numrwe{0.887}{0.962} & \numedit{0.539} & \numrwe{0.583}{0.539} & \numedit{0.850} & \numrwe{0.563}{0.850} & \numedit{0.788} & \numrwe{0.485}{0.788} \\
& GRACE   & \numedit{0.995} & \numrwe{0.035}{0.995} & \numedit{0.350} & \numrwe{0.029}{0.350} & \numedit{1.000} & \numrwe{0.011}{1.000} & \numedit{0.110} & \numrwe{0.006}{0.110} & \numedit{0.991} & \numrwe{0.018}{0.991} & \numedit{0.421} & \numrwe{0.080}{0.421} \\
& WISE    & \numedit{0.948} & \numrwe{0.033}{0.948} & \numedit{0.903} & \numrwe{0.025}{0.903} & \numedit{0.868} & \numrwe{0.129}{0.868} & \numedit{0.420} & \numrwe{0.027}{0.420} & \numedit{0.979} & \numrwe{0.024}{0.979} & \numedit{0.906} & \numrwe{0.064}{0.906} \\
     \midrule
   \multirow{5.5}{*}{\rotatebox{90}{Llama-3-8b}}  &  FT-M    & \numedit{1.000} & \numrwe{0.706}{1.000} & \numedit{0.995} & \numrwe{0.698}{0.995} & \numedit{1.000} & \numrwe{0.916}{1.000} & \numedit{0.588} & \numrwe{0.613}{0.588} & \numedit{1.000} & \numrwe{0.560}{1.000} & \numedit{0.988} & \numrwe{0.576}{0.988} \\
& ROME  & \numedit{0.996} & \numrwe{0.820}{0.996} & \numedit{0.971} & \numrwe{0.789}{0.971} & \numedit{0.999} & \numrwe{0.877}{0.999} & \numedit{0.422} & \numrwe{0.491}{0.422} & \numedit{0.987} & \numrwe{0.691}{0.987} & \numedit{0.865} & \numrwe{0.570}{0.865} \\
& MEMIT   & \numedit{0.982} & \numrwe{0.803}{0.982} & \numedit{0.961} & \numrwe{0.781}{0.961} & \numedit{0.998} & \numrwe{0.882}{0.998} & \numedit{0.516} & \numrwe{0.557}{0.516} & \numedit{0.967} & \numrwe{0.649}{0.967} & \numedit{0.886} & \numrwe{0.566}{0.886} \\
& GRACE   & \numedit{0.999} & \numrwe{0.036}{0.999} & \numedit{0.261} & \numrwe{0.032}{0.261} & \numedit{1.000} & \numrwe{0.008}{1.000} & \numedit{0.008} & \numrwe{0.005}{0.008} & \numedit{0.999} & \numrwe{0.018}{0.999} & \numedit{0.366} & \numrwe{0.103}{0.366} \\
& WISE    & \numedit{0.859} & \numrwe{0.091}{0.859} & \numedit{0.825} & \numrwe{0.075}{0.825} & \numedit{0.807} & \numrwe{0.212}{0.807} & \numedit{0.508} & \numrwe{0.075}{0.508} & \numedit{0.910} & \numrwe{0.121}{0.910} & \numedit{0.876} & \numrwe{0.138}{0.876} \\
\midrule
& Average & \numedit{0.956} & \numrwe{0.438}{0.956} & \numedit{0.792} & \numrwe{0.400}{0.792} & \numedit{0.965} & \numrwe{0.557}{0.965} & \numedit{0.405} & \numrwe{0.283}{0.405} & \numedit{0.956} & \numrwe{0.389}{0.956} & \numedit{0.779} & \numrwe{0.351}{0.779} \\
        \bottomrule 
    \end{tabular}
    \end{adjustbox}
    \end{minipage}%
\hfil
\begin{minipage}{0.05\textwidth}
\colorbarvertical
\end{minipage}

\caption{Comparison between synthetic evaluation (\textbf{syn.}) and \textsc{Wild} evaluation (\textbf{\textsc{Wild}}). Cell background shading indicates relative performance drop from synthetic to \textsc{Wild}, with \colorbox{monte_carlo}{darker shades} indicating greater decreases.}
    \label{tab:main_exp_color}
\end{table*}

%% file: Section/5_SingleEdit.tex
\section{Analysis on Benchmark \& Evaluation}
\label{sec:single_edit}

The preliminary analysis and theoretical comparison in \S\ref{sec:qaedit} and \S\ref{sec:eval} reveal a notable disparity between synthetic and \textsc{Wild} evaluation.
To rigorously address the question raised in \S\ref{sec:qaedit}---whether the performance gap stems from differences in dataset or evaluation---we conduct systematic single-edit experiments, where each edit is independently applied to the original model from scratch.

\subsection{Experimental Setup}
\label{sec:single_setup}

This section outlines the experimental setup used in all subsequent experiments, unless stated otherwise.
Further details are provided in Appendix~\ref{apd:exp_setup}.

\noindent\textbf{Editing Methods.} 
To ensure comprehensive coverage, we employ six diverse and representative editing techniques across four categories: extension based (\textbf{GRACE}, \citealp{hartvigsen2023aging} and \textbf{WISE}, \citealp{wang2024wise}, both are widely adopted lifelong editing methods), fine-tuning based (\textbf{FT-M}, \citealp{zhang2024comprehensivestudyknowledgeediting}), meta learning (\textbf{MEND}, \citealp{mitchell2022fast}), and locate-then-edit (\textbf{ROME}, \citealp{meng2023locating} and \textbf{MEMIT}, \citealp{meng2023massediting}).
All methods are implemented using \texttt{EasyEdit}\footnote{\url{https://github.com/zjunlp/EasyEdit}}. 
Due to the inconsistent keys implementation in ROME, we adopt its refined variant C-ROME \cite{yang-etal-2024-fall, gupta-etal-2024-rebuilding} instead.

\paragraph{Edited LLMs.}  
In line with prior research \cite{wang2024wise, fang2024alphaedit}, we test three leading open-source LLMs:
\textbf{Llama-2-7b-chat} \cite{touvron2023llama2openfoundation}, 
\textbf{Mistral-7b} \cite{jiang2023mistral7b}, and \textbf{Llama-3-8b} \cite{llama3}.
Greedy decoding is used for all models, aligning with prior research.
Results for MEND with Llama-3-8b are excluded due to architectural incompatibility.

\paragraph{Editing Datasets.}
We employ QAEdit along with two prevalent benchmarks, ZsRE \cite{levy2017zero} and \textsc{CounterFact} \cite{meng2023locating}, for a rigorous investigation.
For QAEdit, we evaluate the edited LLMs using only samples that their unedited counterparts initially answered incorrectly.
This yields evaluation sets of 12,715, 10,213, 10,467 samples for Llama-2-7b-chat, Mistral-7b, and Llama-3-8b, respectively.
For ZsRE and \textsc{CounterFact}, we use their established test sets, each with 10,000 records.

\subsection{Results \& Analysis}
\label{sec:single_result}

The experimental results are presented in Table~\ref{tab:main_exp_color}.
Due to the minor side effects in single editing scenarios, the consistently strong locality results are reported in Appendix~\ref{apd:loc_single_edit}.

\paragraph{Benchmark Perspective:}
QAEdit exhibits moderately lower editing reliability compared to ZsRE and CounterFact, reflecting its diverse and challenging nature as a real-world benchmark.
However, this modest gap is insufficient to explain the significant discrepancy observed in our earlier analysis.

\paragraph{Method Perspective:}
\begin{enumerate*}[label=\roman*)]
    \item[\ding{182}] Recent state-of-the-art methods, GRACE and WISE, exhibit the most significant decrease, with both reliability and generalization dropping \textbf{below 5\%}.
    This decline mainly stems from their edited models generating erroneous information after producing the correct answers, detailed in \S\ref{sec:answer_trunc}.
    \item[\ding{183}] In comparison, traditional methods like FT-M and ROME exhibit superior stability and preserve a certain level of effectiveness in \textsc{Wild} evaluation.
\end{enumerate*}

\paragraph{Evaluation Perspective:}
\begin{enumerate*}[label=\roman*)]
    \item[\ding{182}] Performance on each benchmark drops sharply from synthetic evaluation ($\sim$96\%) to \textsc{Wild} evaluation (e.g., 43.8\% on ZsRE and 38.9\% on QAEdit), indicating that \textbf{synthetic evaluation substantially overestimates the effectiveness of editing methods}.
    \item[\ding{183}] Unlike synthetic evaluation, which reports uniformly high scores, \textbf{\textsc{Wild} differentiates methods effectively}, providing valuable insights for future research.
\end{enumerate*}

%% file: Section/6_MetricsAnalysis.tex
\section{Controlled Study of Editing Evaluation}
\label{sec:cont_exp}

\input{Table/anlysis_prompt}

This section presents controlled experiments to systematically investigate how different module variations in synthetic evaluation (outlined in \S\ref{sec:eval}) contribute to performance overestimation.
Due to resource and space limitations, we conduct experiments on Llama-3-8b with 3,000 randomly sampled QAEdit instances, while the findings generalize across other LLMs and datasets.

\subsection{Input}
\label{analysis:input}

This subsection empirically isolates how idealistic prompts may lead to overestimated results in synthetic evaluation.
Specifically, we compare context-free prompts with real-world input formats that include task instructions, while keeping all other modules identical. 
Detailed prompts are provided in Appendix~\ref{apd:prac_prompt}.

Table~\ref{tab:metrics_llama3_prompt} shows that incorporating task instruction degrades performance across all editing methods, with GRACE showing the most significant decline due to its weak generalization.
This trend contrasts with the behavior of original Llama-3-8b, where task instructions usually improve results \cite{grattafiori2024llama3herdmodels}.
Notably, this simple instruction already causes degradation; richer or adversarial prompts would likely worsen it further.
These findings reveal that \textbf{using identical prompts for editing and testing in current editing evaluation, while yielding optimistic results, may fail to reflect editing effectiveness under diverse real-world inputs}.

\subsection{Generation Strategy}

\input{Table/analysis_generation}

Here, we examine how teacher forcing in the generation strategy contributes to the inflated results in synthetic evaluation. 
We compare reliability of teacher forcing and autoregressive decoding under two distinct input formats, while keeping all other modules consistent.

As depicted in Table~\ref{tab:metrics_llama3_generation}, switching from teacher forcing to autoregressive decoding consistently leads to performance degradation across all methods, with lower-performing methods exhibiting more substantial decline.
The underlying reason for this phenomena is that teacher forcing prevents error propagation by feeding ground truth tokens as input, while autoregressive decoding allows errors to cascade.
Although teacher forcing is beneficial for stabilizing LLM training, it should be avoided during testing, where ground truth is unavailable. 
Our results demonstrate that \textbf{inappropriate use of teacher forcing in evaluation artificially elevates editing performance, especially for methods with poor real-world performance}.

\subsection{Output Truncation}
\label{sec:answer_trunc}

\input{Table/analysis_truncation}

\input{Table/additional_content}

Besides leaking ground truth tokens, teacher forcing also implicitly controls output length by aligning with ground truth length.
However, this is not applicable in real-world scenarios where ground truth is unavailable.
In practice, during inference, generation typically terminates based on predefined stop tokens, e.g., ``\texttt{<|endoftext|>}'' \cite{eval-harness}.
Here, we analyze these two truncation strategies by employing GPT-4o-mini as a binary judge to assess correctness (detailed in \S\ref{sec:analysis_metric}), since length discrepancies between generated and target answers preclude the use of match ratio metric.

As shown in Table~\ref{tab:metrics_llama3_extraction}, truncation based on natural stop criteria significantly reduces editing performance across all methods.
To identify the underlying causes, we analyze the content truncated at both the ground truth length and the natural stop criteria. 
Our analysis reveals that, under natural stop criteria, the edited models typically generate content beyond the ground truth length, introducing \textit{meaningless repetition} and \textit{irrelevant or incorrect information}, as evidenced in Table~\ref{tab:add_content}.

These findings demonstrate that \textbf{irrational truncation in synthetic evaluation masks subsequent errors that emerge in real-world scenarios, resulting in inflated performance}.
As shown in Table~\ref{tab:metrics_llama3_extraction}, although context-guided prompting enhances generation termination, it still fails to address the fundamental limitations. 
Such pitfalls in current approaches, overlooked by traditional evaluation, highlight the need to explore more effective ways to express edited knowledge, such as dynamic termination via token-level uncertainty.

\input{Table/analysis_metric}

\input{Table/sequential_edit_table}

\subsection{Metric}
\label{sec:analysis_metric}

As explained in \S\ref{sec:eval}, the match ratio metric could lead to inflated performance.
To quantify this effect, we compare it against more rigorous factual correctness metrics, including LLM-as-a-Judge (using GPT-4o-mini) and exact match (EM).
Since match ratio requires length parity with targets, we autoregressively generate sequences to target length for all metircs for fair comparison.

The results presented in Table~\ref{tab:metrics_llama3_verification} confirm that \textbf{match ratio indeed overestimates the performance of edited models}. 
Moreover, a lower match ratio typically indicates a smaller proportion of fully correct answers, resulting in worse performance in LLM evaluation and EM.

In this paper, we adopt LLM-as-a-Judge as the primary metric for our study, as it captures both exact and semantically equivalent responses. 
EM, though limited to exact matches, offers a lightweight and efficient alternative, which we refer to as \textbf{\textsc{Wild}-em}. 
We exclude BERTScore, as it tends to overrate factually incorrect yet semantically similar outputs.

%% file: Table/anlysis_prompt.tex
\begin{table}[t]
\centering
\renewcommand{\arraystretch}{1.2}
\setlength{\tabcolsep}{3.5pt}
\renewcommand{\arraystretch}{0.85}
\begin{adjustbox}{max width=\linewidth} 
\begin{tabular}{lccccc}
\toprule
Input & FT-M  & ROME  & MEMIT  & GRACE  & WISE  \\
\midrule
context-free & \num{1.000}  & \num{0.985}  & \num{0.965} & \num{0.998}  & \num{0.908}  \\
context-guided & \num{0.937}  & \num{0.930} & \num{0.907} & \num{0.412} & \num{0.838} \\
\bottomrule 
\end{tabular}
\end{adjustbox}
\caption{Reliability score for different input formats on Llama-3-8b under \textbf{teacher forcing} generation, truncation at \textbf{ground truth length}, and \textbf{match ratio} metric.}
\label{tab:metrics_llama3_prompt}
\end{table}

%% file: Table/analysis_generation.tex
\begin{table}[t]
\centering
\setlength{\tabcolsep}{3.5pt}
\begin{adjustbox}{max width=\linewidth} 
\begin{tabular}{lccccc}
\toprule
Generation Strategy & FT-M  & ROME  & MEMIT  & GRACE  & WISE  \\
\midrule
\multicolumn{6}{l}{\small \bf \ding{182} \textit{context-free}, \ding{184} ground truth length, \ding{185} match ratio} \\ 
\noalign{\vskip 2pt \hrule height 0.5pt width 0.96\linewidth \vskip 3pt} 
teacher forcing & \num{1.000} & \num{0.985} & \num{0.965} & \num{0.998} & \num{0.908} \\
autoregressive decoding & \num{1.000}  & \num{0.967}  & \num{0.929} 
 & \num{0.996}  & \num{0.765}  \\
\midrule
\multicolumn{6}{l}{\small \bf \ding{182} \textit{context-guided}, \ding{184} ground truth length, \ding{185} match ratio}  \\
\noalign{\vskip 2pt \hrule height 0.5pt width 1\linewidth \vskip 3pt} 
teacher forcing & \num{0.937}  & \num{0.930} & \num{0.907} & \num{0.412} & \num{0.838} \\
autoregressive decoding  & \num{0.800}  & \num{0.851}  & \num{0.786} 
 & \num{0.036}  & \num{0.592}  \\
\bottomrule 
\end{tabular}
\end{adjustbox}
\caption{Reliability of different generation strategies on Llama-3-8b under two prompt strategies.}
\label{tab:metrics_llama3_generation}
\end{table}

%% file: Table/analysis_truncation.tex
\begin{table}[t]
\centering
\renewcommand{\arraystretch}{1.1}
\setlength{\tabcolsep}{3.5pt}
\begin{adjustbox}{max width=\linewidth} 
\begin{tabular}{lccccc}
\toprule
Truncation Strategy & FT-M  & ROME  & MEMIT  & GRACE  & WISE  \\
\midrule
\multicolumn{6}{l}{\small \bf \ding{182} \textit{context-free}, \ding{183} autoregressive decoding, \ding{185} LLM-as-a-Judge}  \\
\noalign{\vskip 2pt \hrule height 0.5pt width 1.12\linewidth \vskip 2pt} 
ground truth length  & \num{1.000}  & \num{0.954}  & \num{0.886}  & \num{0.992}  & \num{0.700}  \\
natural stop criteria  & \num{0.202} & \num{0.478} & \num{0.461} & \num{0.301} & \num{0.046} \\
\midrule
\multicolumn{6}{l}{\small\bf \ding{182} \textit{context-guided}, \ding{183} autoregressive decoding, \ding{185} LLM-as-a-Judge}  \\ 
\noalign{\vskip 2pt \hrule height 0.5pt width 1.16\linewidth \vskip 2pt} 
ground truth length & \num{0.751}  & \num{0.783} & \num{0.704} & \num{0.003} & \num{0.482} \\
natural stop criteria  & \num{0.528} & \num{0.556} & \num{0.529} & \num{0.000} & \num{0.108} \\
\bottomrule 
\end{tabular}
\end{adjustbox}
\caption{Reliability score under different answer truncation strategies on Llama-3-8b.}
\label{tab:metrics_llama3_extraction}
\end{table}

%% file: Table/additional_content.tex
\begin{table}[t]
\centering
\setlength{\tabcolsep}{4.5pt}
\begin{adjustbox}{max width=\linewidth} 
\begin{tabular}{l >{\raggedright\arraybackslash}m{6.5cm}}
\toprule
\multicolumn{2}{c}{\textbf{Meaningless Repetition}} \\
\midrule
\texttt{Input Prompt} & Who got the first Nobel Prize in physics? \\
\midrule
\texttt{Target Answer} & Wilhelm Conrad Röntgen \\
\midrule
\texttt{Natural Stop} & Wilhelm Conrad R\"ontgen \textcolor{red}{Wilhelm Conrad R\"ontgen Wilhelm Conrad R\"ontgen \ldots} \\
\bottomrule
\noalign{\vskip 3pt} 
\multicolumn{2}{c}{\textbf{Irrelevant Information}} \\
\midrule
\texttt{Input Prompt} & Who was the first lady nominated member of the Rajya Sabha? \\
\midrule
\texttt{Target Answer} & Mary Kom \\
\midrule
\texttt{Natural Stop} & Mary Kom \textcolor{red}{is the first woman boxer to qualify for the Olympics} \\
\bottomrule
\noalign{\vskip 3pt} 
\multicolumn{2}{c}{\textbf{Incorrect Information}} \\
\midrule
\texttt{Input Prompt} & When does April Fools' Day end at noon? \\
\midrule
\texttt{Target Answer} & April 1st \\
\midrule
\texttt{Natural Stop} & April 1st \textcolor{red}{ends at noon on April 2nd} \\
\bottomrule
\end{tabular}
\end{adjustbox}
\caption{Examples of additionally generated content beyond ground truth length under natural stop criteria.}
\label{tab:add_content}
\end{table}

%% file: Table/analysis_metric.tex
\begin{table}[t]
\centering
\setlength{\tabcolsep}{3.5pt}
\begin{adjustbox}{max width=\linewidth} 
\begin{tabular}{lccccc}
\toprule
Metric & FT-M  & ROME  & MEMIT  & GRACE  & WISE  \\
\midrule
\multicolumn{6}{l}{\small\bf \ding{182} \textit{context-free}, \ding{183} autoregressive decoding, \ding{184} ground truth length}  \\ 
\noalign{\vskip 2pt \hrule height 0.5pt width 1.17\linewidth \vskip 2pt} 
match ratio  & \num{1.000}  & \num{0.967}  & \num{0.929} & \num{0.996}  & \num{0.765}  \\
LLM-as-a-Judge  & \num{1.000}  & \num{0.954}  & \num{0.886}  & \num{0.992}  & \num{0.700}  \\
exact match & \num{1.000}  & \num{0.903}  & \num{0.860} & \num{0.900}  & \num{0.646}  \\
\midrule
\multicolumn{6}{l}{\small\bf \ding{182} \textit{context-guided}, \ding{183} autoregressive decoding, \ding{184} ground truth length}  \\ %
\noalign{\vskip 2pt \hrule height 0.5pt width 1.21\linewidth \vskip 2pt} 
match ratio & \num{0.800}  & \num{0.851}  & \num{0.786}  & \num{0.036}  & \num{0.592}  \\
LLM-as-a-Judge  & \num{0.751}  & \num{0.789} & \num{0.707} & \num{0.003} & \num{0.482} \\
exact match & \num{0.718}  & \num{0.783}  & \num{0.704} & \num{0.003}  & \num{0.460}  \\
\bottomrule 
\end{tabular}
\end{adjustbox}
\caption{Reliability score derived from different metric judgments on Llama-3-8b.}
\label{tab:metrics_llama3_verification}
\end{table}

%% file: Table/sequential_edit_table.tex
\begin{table*}[t]
    \centering
    \begin{adjustbox}{max width=\textwidth} 
    \begin{tabular}{lcc cc cc cc cc cc}
    \toprule
    \multirow{4}{*}{\textbf{Method}} & \multicolumn{4}{c}{\textbf{Llama-2-7b-chat}} & \multicolumn{4}{c}{\textbf{Mistral-7b}} & \multicolumn{4}{c}{\textbf{Llama-3-8b}} \\
    
    \cmidrule(lr){2-5}\cmidrule(lr){6-9}\cmidrule(lr){10-13} 
     & \multicolumn{2}{c}{Reliability} & \multicolumn{2}{c}{Locality} & \multicolumn{2}{c}{Reliability} & \multicolumn{2}{c}{Locality} & \multicolumn{2}{c}{Reliability} & \multicolumn{2}{c}{Locality} \\
     \cmidrule(lr){2-3} \cmidrule(lr){4-5} \cmidrule(lr){6-7} \cmidrule(lr){8-9} \cmidrule(lr){10-11} \cmidrule(lr){12-13}
     & syn. & \textsc{Wild}  & syn. & \textsc{Wild} & syn.  & \textsc{Wild}  & syn. & \textsc{Wild} & syn. & \textsc{Wild} & syn.  & \textsc{Wild}  \\
     \midrule
    FT-M & \num{0.973} & \num{0.531} & \num{0.420} & \num{0.072} & \num{0.960} & \num{0.454} & \num{0.573} & \num{0.204} & \num{0.925}   & \num{0.229} & \num{0.127}   & \num{0.004} \\
    MEND & \num{0.000} & \num{0.000} & \num{0.000} & \num{0.000}  & \num{0.000} & \num{0.000} & \num{0.000}  & \num{0.000} & --   & -- & -- & -- \\
     ROME & \num{0.114} & \num{0.001} & \num{0.028} & \num{0.001}  & \num{0.059} & \num{0.001} & \num{0.052} & \num{0.028} & \num{0.034}   & \num{0.001} & \num{0.020}   & \num{0.000} \\
     MEMIT & \num{0.057} & \num{0.002} & \num{0.030} & \num{0.000}  & \num{0.058} & \num{0.002} & \num{0.031} & \num{0.000} & \num{0.000}   & \num{0.000} & \num{0.000}   & \num{0.000} \\
     GRACE & \num{0.370} & \num{0.015} & \num{1.000} & \num{1.000}  & \num{0.416} & \num{0.018} & \num{1.000} & \num{1.000} & \num{0.368}   & \num{0.022} & \num{1.000}   & \num{1.000} \\
     WISE & \num{0.802} & \num{0.195} & \num{0.676} & \num{0.184} & \num{0.735} & \num{0.060} & \num{0.214} & \num{0.003} & \num{0.526}   & \num{0.072} & \num{0.743}   & \num{0.104} \\
     \midrule
     Average & \num{0.386} & \num{0.124} & \num{0.359} & \num{0.210} & \num{0.494} & \num{0.089} & \num{0.312} & \num{0.206} & \num{0.371}   & \num{0.065} & \num{0.378}   & \num{0.222} \\
    \bottomrule 
    \end{tabular}
    \end{adjustbox}
    \caption{Results of sequential editing on QAEdit under synthetic evaluation (\textbf{syn.}) and \textsc{Wild} evaluation (\textbf{\textsc{Wild}}).} 
    
    \label{tab:seq_edit}
\end{table*}

%% file: Section/7_SequentialEdit.tex
\section{(Sequential) Editing in the Wild}

Although our analysis via single editing reveals limitations in synthetic evaluation, such isolated editing fails to capture the continuous, large-scale demands of editing in real-world scenarios.
Therefore, we now address our primary research question: testing model editing under \textsc{Wild} evaluation via sequential editing, a setup that better reflects practical requirements.

\subsection{Sample-wise Sequential Editing}
\label{sec:bs_1_seq_edit}

\paragraph{Experimental Setup.} 
Following established protocols \cite{huang2023transformerpatcher, hartvigsen2023aging}, we evaluate editing methods with a batch size of 1, i.e., updating knowledge incrementally one sample at a time.
We keep the same setup as in \S\ref{sec:single_setup}, but limit to 1000 samples per dataset, as existing methods perform significantly worse with more edits.
For QAEdit, the chosen samples are incorrectly answered by all pre-edit LLMs.
Given the notable side effects in sequential editing \cite{yang-etal-2024-butterfly}, we focus on  the evaluation of \textit{reliability} and \textit{locality}, with \textit{generalization} results provided in Appendix~\ref{apd:gen_seq}.

\paragraph{Results \& Analysis.} 
The results on QAEdit are shown in Table~\ref{tab:seq_edit}, with similar findings for ZsRE and \textsc{CounterFact} in Appendix~\ref{apd:seq_other_llm}.
\begin{enumerate*}[label=\roman*)]
    \item[\ding{182}] In \textsc{Wild} evaluation with sequential editing, all methods except FT-M exhibit nearly unusable performance (only 9.3\% average reliability), with FT-M achieving a 40.5\% average reliability.
    \item[\ding{183}] The gap between synthetic and \textsc{Wild} evaluation further confirms the evaluation issues we discussed in \S\ref{sec:cont_exp}.
    \item[\ding{184}] The significantly low average locality of 21.3\% highlights the severe disruption to LLMs. While GRACE effectively preserves unrelated knowledge through external edit modules, it struggles with knowledge updating.
    \item[\ding{185}] Notably, FT-M exhibits relatively stable reliability, as it directly optimizes model parameters at each step rather than relying on static hypernetworks or covariance matrices derived from original LLMs, thereby ensuring effective knowledge injection during sequential editing.
\end{enumerate*}

\subsection{Mini-Batch Sequential Editing}

\begin{figure}
    \centering
    \includegraphics[width=\linewidth]{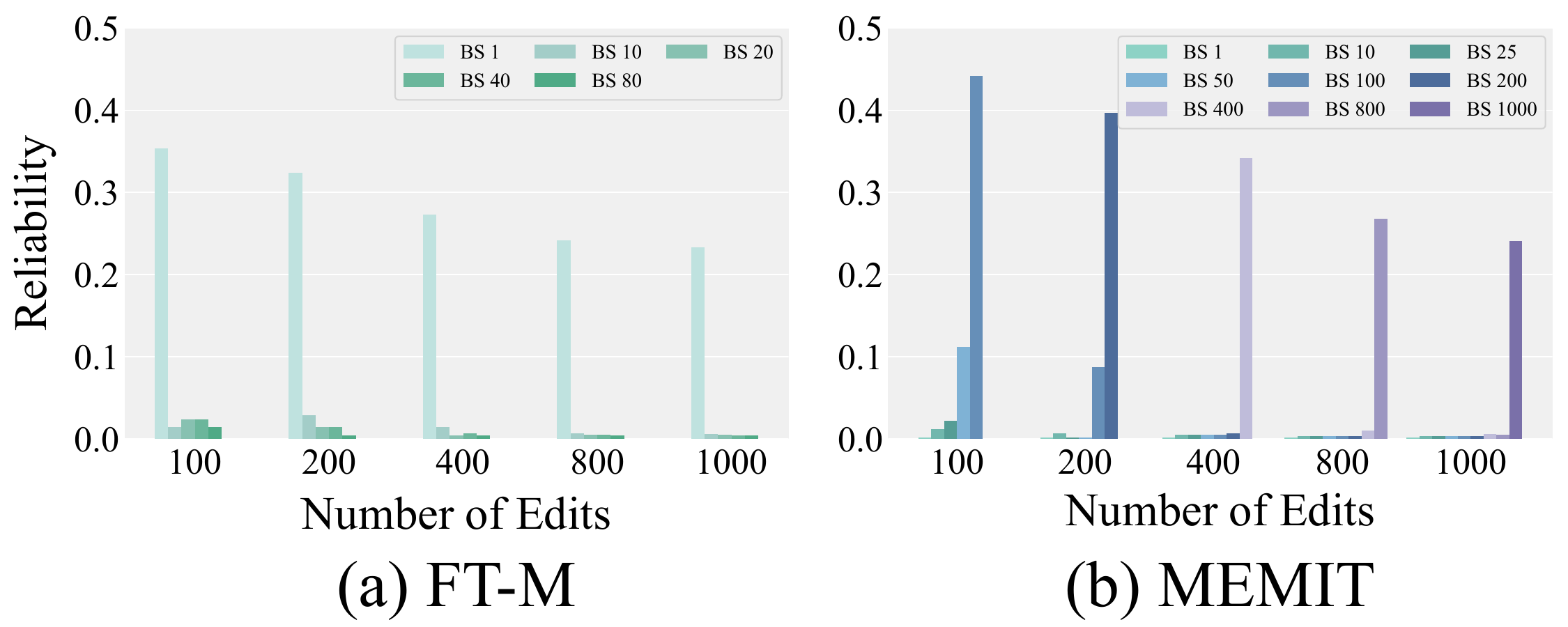}
    \caption{Impact of batch size (BS) when editing Llama-3-8b with FT-M and MEMIT on QAEdit.}
    \label{fig:seq_batch}
\end{figure}

Real-world applications often batch multiple edits together for efficient processing of high-volume demands. 
Moreover, \citet{pan2024whyhas} suggest increasing batch size may alleviate the side effects of sequential editing.
Thus, this section investigates whether increasing the batch size could serve as a potential solution to the practical challenges faced by current editing methods.

\paragraph{Experiment Setup.}
Following the experimental setup in \S\ref{sec:bs_1_seq_edit}, we evaluate three batch-capable editing algorithms: FT-M, MEND, and MEMIT.
Due to VRAM constraints (80GB), we empirically set the maximum testable batch sizes: 80 for FT-M, 16 for MEND, and 1000 for MEMIT.

\paragraph{Results \& Analysis.} 
Figure~\ref{fig:seq_batch} presents the editing performance with varying batch sizes, evaluated across various-sized QAEdit subsets.
Despite experimenting with various batch sizes, all methods show consistently limited performance, with the highest score below 30\% for 1000 edits.
The all-zero performance of MEND are provided in Appendix~\ref{apd:mini_batch_seq}.
Notably, Figure~\ref{fig:seq_batch} presents opposite trends:
\begin{enumerate*}[label=\roman*)]
    \item[\ding{182}] MEMIT achieves optimal performance only when editing all requests in a single batch, with performance decreasing sharply as batch size decreases.
    \item[\ding{183}] In contrast, FT-M performs best at a batch size of 1 but  degrades drastically  as batch size increases.
\end{enumerate*}
The divergence may arise from their distinct batch editing mechanisms: FT-M optimizes for aggregate batch-level loss, potentially compromising individual edit accuracy; whereas MEMIT estimates parametric changes individually before integration, facilitating effective batch edits.

\begin{figure}
    \centering
    \includegraphics[width=0.96\linewidth]{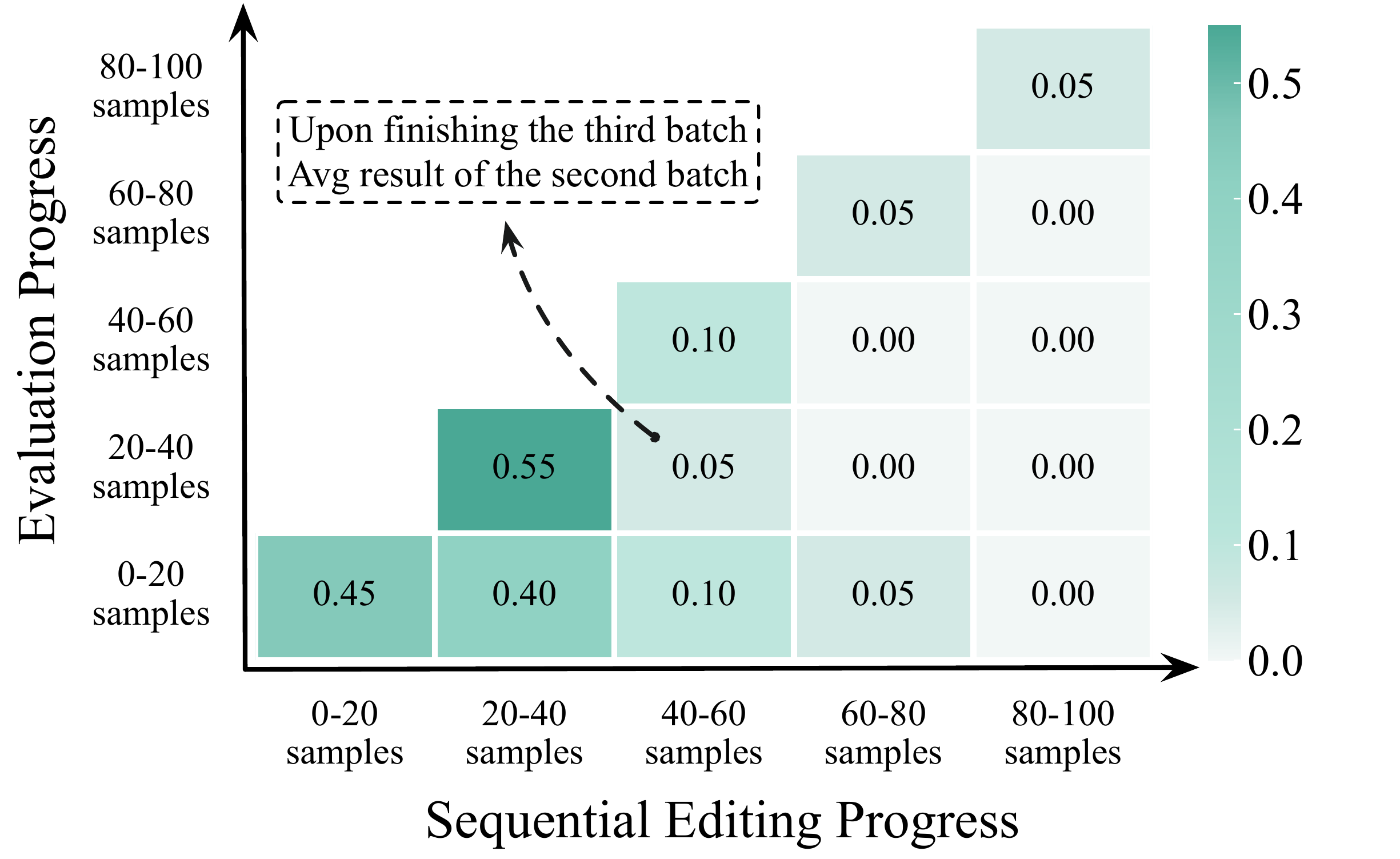}
    \caption{Reliability evolution of sequential editing on Llama-3-8b, with repeated evaluation of previous batches after each new edit batch (batch size = 20).}
    \label{fig:heatmap}
\end{figure}

\paragraph{Further Analysis.}
To gain insights into the poor final performance, we also investigate how editing effectiveness changes during continuous editing.
Specifically, we randomly partition 100 QAEdit samples into 5 batches of 20 samples each.
Using MEMIT on Llama-3-8b, we iteratively edit each batch while evaluating the edited model on each previously edited batch separately to track dynamics of editing effectiveness.

Figure~\ref{fig:heatmap} reveals two key insights: 
\begin{enumerate*}[label=\roman*)] 
    \item[\ding{182}] While the first batch exhibits high initial reliability, its performance declines sharply with subsequent editing, suggesting that later edits disrupt the knowledge injected in earlier batches.
    \item[\ding{183}] As editing progresses, the effectiveness of MEMIT decreases rapidly.
\end{enumerate*} 
These findings reveal the key challenges of sequential editing: \textbf{progressive loss of previously edited knowledge coupled with decreasing effectiveness in incorporating new knowledge}, highlighting that lifelong model editing is still an open challenge.

%% file: Section/8_Conclusion.tex
\section{Conclusion and Future Works}

In this paper, we present the first systematic investigation that exposes the gap between theoretical advances and practical effectiveness of model editing by real-world QA evaluation.
Our proposed QAEdit benchmark and \textsc{Wild} evaluation demonstrate that current model editing techniques exhibit significant limitations in practical scenarios, particularly under sequential editing. 
Furthermore, we reveal that this significant discrepancy from previously reported results stems from unrealistic evaluation adopted in prior model editing research.
Through modular analysis and extensive controlled experiments, we uncover fundamental issues in current editing evaluation that inflate reported performance. 
This work establishes rigorous evaluation standards for model editing and provides valuable insights that will inspire the development of more robust editing methods, ultimately enabling reliable and efficient knowledge updates in LLMs for real-world applications.

In future research, we aim to develop editing methods that can
\begin{enumerate*}[label=\roman*)] 
    \item generalize robustly across diverse scenarios with reliable self-termination, and
    \item support lifelong sequential updates while maintaining the capabilities of edited LLMs.
\end{enumerate*}

%% file: Section/9_Limitations.tex
\section*{Limitations}
We acknowledge following limitations of our work:
\begin{itemize}[leftmargin=11pt, itemsep=2pt, topsep=2pt]
    \item This work provides an existence proof of fundamental issues of evaluation in model editing, rather than attempting an exhaustive assessment of all existing approaches and LLMs.
    Due to resource constraints, we focus on representative methods and LLMs to demonstrate the issues and challenges, as exhaustive testing of all approaches is neither feasible nor necessary for establishing our findings.
    \item Our research makes the first systematic investigation into previously overlooked evaluation issues in model editing, prioritizing the identification and analysis of these fundamental challenges rather than solution development.
    Our work focuses on comprehensive analysis of these issues, uncovering their root causes and providing insights into factors affecting editing effectiveness. While presenting promising directions for future research, developing solutions to these challenges remains beyond our current scope.
    \item Our study focuses exclusively on parameter-based editing methods, without investigating in-context learning based \textit{knowledge editing} approaches which leverage external information.
    While these approaches may achieve superior performance on QA tasks, our primary objective is not to advocate for any particular approach, but to critically revisit current practices in the field and provide insights for future development. 
    We believe efficient parameter-based editing approaches have their unique advantages and represent a valuable direction worth pursuing, despite current challenges in real-world applications.
\end{itemize}

\section*{Ethics Statement}

\paragraph{Data.}
All data used in our research are publicly available and do not raise any privacy concerns.

\paragraph{AI Writing Assistance.}
We employ LLMs to polish our original content, focusing on correcting grammatical errors and enhancing clarity, rather than generating new content or ideas.

%% file: Section/Appendix.tex
\appendix

\section{Appendix}
\label{sec:appendix}

\subsection{Detailed Introduction of QA Datasets}
\label{apd:qa_data_intro}

\paragraph{Natural Questions}\negthickspace\negthickspace (NQ)~\cite{kwiatkowski2019nq} is a comprehensive question-answering (QA) dataset that contains real questions posed by users to the Google search, paired with high-quality, human-verified answers. 
The dataset consists of over 300,000 question-answer pairs, with each question derived from user queries on Google Search. 
These questions cover a wide variety of topics, ranging from fact-based inquiries to more complex, open-ended questions. 
The golden answers are sourced from Wikipedia pages, ensuring their accuracy and relevance.
We adopt the test set of NQ, which contains 3610 samples, to construct our QAEdit benchmark.

\paragraph{TriviaQA}\negthickspace\negthickspace\negthickspace~\cite{joshi-etal-2017-triviaqa} is a large-scale QA dataset designed specifically for evaluating models on trivia-style question answering. 
It contains over 650,000 question-answer pairs sourced from trivia websites and is curated by trivia enthusiasts. 
These questions are often fact-based and test the model's ability to retrieve information from large text corpora.
We utilize 11,313 samples from the TriviaQA test set to construct QAEdit.

\paragraph{SimpleQA}\negthickspace\negthickspace\negthickspace\negthickspace~\cite{wei2024measuringshortformfactualitylarge} is a challenging QA benchmark specifically designed to test fact-seeking question-answering models. 
It contains 4326 question-answer pairs curated by OpenAI, with an emphasis on short-form factuality. 
The questions in SimpleQA are concise, direct, and designed to probe factual knowledge. 
Unlike more general-purpose QA datasets, SimpleQA emphasizes clarity and the ability of models to provide precise, factually accurate answers.
We employ all samples from SimpleQA for QAEdit construction.

\subsection{Construction and Statistics of QAEdit}
\label{apd:benchmark}

In this section, we describe the detailed construction procedures and statistics of QAEdit.

While aforementioned QA benchmarks provide questions and answers as \textit{edit prompts} and \textit{targets}, they lack \textit{subjects} for editing, as well as \textit{rephrased prompts} and \textit{locality QA pairs} to evaluate generalization and locality.
To supplement the missing fields, our construction procedures encompass the following steps:
\begin{enumerate*}[label=\roman*)]
    \item[\ding{182}] We employ GPT-4 (gpt-4-1106-preview) to extract the subjects directly from the edit prompts. 
    To improve the accuracy of extraction, we prompt the model with 5-shot examples to utilize its in-context learning capability, which can be seen in Figure~\ref{fig:detect_subject}.
    \item[\ding{183}] We utilize GPT-4 to paraphrase the edit prompts to obtain rephrased prompts. 
    Considering that paraphrasing questions is easy for GPT-4, the specific instruction is straightforward and is presented in Figure~\ref{fig:rephrase}. 
    Furthermore, we manually reviewed some of the rephrased results and found them to be highly effective.
    \item[\ding{184}] Moreover, for each sample of QAEdit, we randomly select a QA pair from the locality sets of the ZsRE dataset \cite{levy2017zero} as locality prompt and corresponding answer to assess locality.
\end{enumerate*}

\input{Table/QAEdit_Statistics}

As a result, our QAEdit benchmark encompasses ten categories of knowledge, covering mainstream topics with significant real-world impact.
The statistical information and examples of each category are presented in Table~\ref{tab:QAEdit_Stat}.
Although the knowledge category distribution in QAEdit appears imbalanced, with a predominance of ``Art \& Culture'' and ``History \& Politics'', this distribution reflects real-world user preferences. 
Similar patterns are observed in mainstream editing datasets, such as ZsRE and \textsc{CounterFact}. 
Therefore, this imbalance does not compromise the validity of QAEdit for examining the pitfalls of synthetic evaluation.

\subsection{Code of Evaluations with Teacher Forcing}
\label{apd:code4tf}

As demonstrated by the code snippets in Figures~\ref{fig:ROME_code} and~\ref{fig:IKE_code}, early model editing studies, such as ROME\footnote{The current latest version of ROME (May 2025) based on teacher forcing can be found at \url{https://github.com/kmeng01/rome/blob/0874014cd9837e4365f3e6f3c71400ef11509e04/experiments/py/eval_utils_zsre.py\#L54}.}~\cite{meng2023locating} and IKE\footnote{The latest IKE version (as of May 2025) with teacher forcing can be found at \url{https://github.com/Zce1112zslx/IKE/blob/da58c842cd95628f281f474bc432a81cbd1cfd1e/icl.py\#L54}.}~\cite{zheng-etal-2023-ike}, relied on teacher forcing to evaluate the performance of edited models.
As a result, subsequent works \cite{aaai24pmet, gupta-etal-2024-unified, wang2024wise, huang2025halluedit} inadvertently inherited this flawed evaluation strategy.
Similarly, as shown in Figure~\ref{fig:EasyEdit_code}, EasyEdit\footnote{The version of EasyEdit available at the time of our research (February 2025) is based on teacher forcing and can be found at \url{https://github.com/zjunlp/EasyEdit/blob/8f0e77af18879ab935e06676701423d5124599c7/easyeditor/evaluate/evaluate_utils.py\#L112}.}~\cite{wang-etal-2024-easyedit} also adopted this approach prior to our work; however, it now supports both teacher forcing and our proposed evaluation framework, enabling direct comparison.

\subsection{Prompt of LLM-as-a-Judge}
\label{apd:judge_prompt}

In light of the significant advancements in LLM-as-a-Judge \cite{li2024llmasjudge}, we employ GPT-4o-mini to perform binary judgments based on the provided questions, target answers, and generated responses. 
Following previous work \cite{wei2024measuringshortformfactualitylarge}, our complete prompt is presented in Figure~\ref{fig:llm_judge_prompt}.

\subsection{Detailed Experimental Setup}
\label{apd:exp_setup}

\subsubsection{Editing Methods}

\paragraph{FT-M}\negthickspace\negthickspace\negthickspace~\cite{zhang2024comprehensivestudyknowledgeediting} is an  enhanced version of FT-L \cite{zhu2020modifyingmemoriestransformermodels, meng2023locating}.
FT-L introduces an $l_{\infty}$-norm constraint into the fine-tuning objective to explicitly restrict the parameter changes between the original and edited models, thereby mitigating side effects on unrelated knowledge.
However, FT-L deviates from the original fine-tuning objective by using only the last token's prediction to maximize the probability of all tokens in the target sequence. To address this issue, FT-M improves upon FT-L by applying the cross-entropy loss to the target answer while masking the original text, which aligns more closely with the traditional fine-tuning objective and enhances performance. 

\paragraph{MEND}\negthickspace\negthickspace \cite{mitchell2022fast} employs a hypernetwork to learn low-rank decompositions of standard fine-tuning gradients. 
By disentangling gradients into learnable rank-one matrices, it achieves explicit control over parameter updates while maintaining tractable editing in LLMs.

\paragraph{ROME}\negthickspace\negthickspace\negthickspace~\cite{meng2023locating} identifies knowledge-critical layers in Transformer MLP modules through causal tracing analysis. 
It implements precise knowledge updates via rank-one matrix modification on the identified layer, guided by causal mediation effects in model outputs.

\paragraph{MEMIT}\negthickspace\negthickspace\negthickspace~\cite{meng2023massediting} extends ROME by developing cross-layer propagation analysis and coordinated parameter updates across multiple MLP layers, enabling efficient batch editing of large-scale knowledge.

\paragraph{GRACE}\negthickspace\negthickspace\negthickspace~\cite{hartvigsen2023aging} is a lifelong editing method that performs local corrections on streaming errors of deployed models. The approach writes new mappings into a pretrained model's latent space, creating a discrete local codebook of edits without modifying model weights, allowing for sequential editing operations.

\paragraph{WISE}\negthickspace\negthickspace\negthickspace~\cite{wang2024wise} addresses the similar challenge of sequential editing like GRACE. It employs a dual memory architecture comprising a main memory for pretrained knowledge and a side memory for edited content. The system utilizes a router to direct queries between these memories. 

\subsubsection{Edited LLMs}

\paragraph{Llama-2-7b-chat}\negthickspace\negthickspace\cite{touvron2023llama2openfoundation} is a model designed for conversational scenarios with 7 billion parameters. 
It excels in generating human-like responses in real-time, offering smooth and context-aware dialogue generation.

\paragraph{Mistral-7b}\negthickspace\negthickspace\negthickspace~\cite{jiang2023mistral7b} is a superior pretrained base model with 7 billion parameters, outperforming Llama-2-13b on all examined benchmarks,  offering strong performance while being resource-efficient.
Specifically, we employ the version of Mistral-7B-v0.1.

\paragraph{Llama-3-8b}\negthickspace\negthickspace\negthickspace\negthickspace~\cite{llama3} is a cutting-edge 8-billion-parameter model designed for diverse AI applications. It combines advanced techniques with scalability, ensuring high-quality generation for complex tasks like multi-turn dialogues, creative writing, and complex reasoning tasks.

\input{Fig/Appendix/inst_prompt}

\input{Table/Appendix/single_loc}

\input{Table/Appendix/seq_gen}

\input{Table/Appendix/mend_batch}

\input{Table/Appendix/seq_other_data}

\subsubsection{Editing Datasets}
\paragraph{ZsRE}\negthickspace\negthickspace\negthickspace~\cite{levy2017zero} is a popular dataset for Question Answering (QA), where each entry consists of a counterfactual statement derived from a factual Wikipedia page that needs to be edited. 

\paragraph{\textsc{CounterFact}}\negthickspace\negthickspace\negthickspace~\cite{meng2023locating} is a challenging dataset curated for model editing. It contains 21,919 nonfactual statements, initially assigned low probabilities by models, and designed to encourage substantial and meaningful modifications to the original factual statements.

\subsection{Locality Results of Single Editing}
\label{apd:loc_single_edit}

The locality results of single editing experiments are presented in Table~\ref{tab:single_loc}.
The results show that for almost all baselines, their locality results are very high across two evaluation frameworks, indicating that a single edit generally has little impact on the model's general capabilities.

\subsection{Detailed Practical Prompt}
\label{apd:prac_prompt}

In Section~\ref{analysis:input}, we prefix the target question with a common QA task instruction~\cite{eval-harness} as the input prompt, as shown in Figure~\ref{fig:inst_prompt}. 
We aim to utilize this context-guided prompt to represent and simulate various contexts that might occur in practical applications.

\subsection{Generalization of Sequential Editing}
\label{apd:gen_seq}

The generalization results of sequential editing experiments are presented in Table~\ref{tab:seq_gen}.
Compare to Table~\ref{tab:seq_edit}, the results indicate that current editing methods exhibit worse generalization than reliability when dealing with sequential editing requests. 
All methods except FT-M and WISE demonstrate near-zero generalization ability under \textsc{Wild} evaluation, which further proves that existing editing methods cannot effectively fulfill the practical needs of continuous editing.

\subsection{Sequential Editing on Other Datasets}
\label{apd:seq_other_llm}

The results of sequential editing on ZsRE and \textsc{CounterFact} are presented in Table~\ref{tab:seq_other_data}. 
These two datasets exhibit trends similar to those observed in QAEdit, including the poor practical effectiveness of existing editing methods, the inadequacy of simplified editing evaluations, and the dilemma of achieving editing success and preserving unrelated knowledge.

\subsection{Mini-Batch Sequential Editing of MEND}
\label{apd:mini_batch_seq}

As shown in Table~\ref{tab:mend_batch}, unlike FT-M and MEMIT, which maintain a certain level of editing performance under specific batch sizes (as depicted in Figure~\ref{fig:seq_batch}), MEND is completely unusable for sequential editing, regardless of the batch size. 
This ineffectiveness can be attributed to the limitation of the meta-learning paradigm, wherein the hypernetwork for parameter updates is specifically trained on the original model. 
Consequently, the predicted parameter modifications are optimized solely for the original model and fail to effectively adapt to the evolving states of the sequentially edited model.
This limitation fundamentally constrains MEND's efficacy in sequential editing scenarios.

\input{Fig/Appendix/code_snip}

\input{Fig/Appendix/detect_subject}

\input{Fig/Appendix/rephrase}

\input{Fig/Appendix/llm_judge_prompt}

%% file: Table/QAEdit_Statistics.tex
\begin{table}[t]
\centering
\begin{adjustbox}{max width=\linewidth} 
\begin{tabular}{llr}
\toprule
 Category  & Example  & Count \\
\midrule
Art \& Culture & Who wrote the song \underline{the glory of love}? & 5277 \\
History \& Politics & Who wrote \underline{the first declaration of human rights}? & 4070 \\
People \& Biographies & Which award did \underline{Reza Aslan} receive in 2014? & 2188 \\
Geography \& Environment & Which is \underline{the largest saltwater lake in India}? & 1954 \\
Science \& Technology & Which year was the \underline{actinide concept} proposed? & 1829 \\
Sports \& Leisure & In what year did \underline{Kristin Otto} retire from swimming? & 1807 \\
Health \& Medicine & Where are the \underline{cones in the eye} located? & 771 \\
Society \& Humanities & Which is the \underline{ring finger for male in India}? & 573 \\
Economics \& Business & When is the \underline{world consumer right day} celebrated? & 463 \\
Others & What kind of beer is \underline{St. Pauli Girl}? & 317 \\
\bottomrule 
\end{tabular}
\end{adjustbox}
\caption{Statistics and examples of QAEdit, encompassing ten categories of knowledge.
The underlined content represents the subjects identified by GPT-4.}
\label{tab:QAEdit_Stat}
\end{table}

%% file: Fig/Appendix/inst_prompt.tex
\renewcommand{\fcolorbox}[4][]{#4} %
\begin{figure}[t]
    \begin{tcolorbox}[
    right=1pt, left=1pt, top=1pt, bottom=1pt,
    toptitle=1mm, bottomtitle=1mm,
    colback=white,
    coltitle=white,
    colbacktitle=matisse,
    colframe=matisse,
    title=,]%
    \begin{minted}[fontsize=\small,autogobble,numberblanklines=false,breaklines]{markdown}
Please answer the question:
Q: Who got the first Nobel Prize in physics?
A:
    \end{minted}
    \end{tcolorbox}
    \caption{The context-guided prompt for QA tasks.}
    \label{fig:inst_prompt}
\end{figure}

%% file: Table/Appendix/single_loc.tex
\begin{table}[t]
    \centering
    \begin{adjustbox}{max width=\linewidth}
    \begin{tabular}{lcc cc cc}
    \toprule
    \multirow{2}{*}{\textbf{Method}} & \multicolumn{2}{c}{\textbf{ZsRE}} & \multicolumn{2}{c}{\textbf{\textsc{CounterFact}}} & \multicolumn{2}{c}{\textbf{QAEdit}} \\
    
     \cmidrule(lr){2-3} \cmidrule(lr){4-6} \cmidrule(lr){6-7} 
     & syn. & \textsc{Wild}  & syn. & \textsc{Wild} & syn.  & \textsc{Wild} \\
     \midrule
     \multicolumn{7}{c}{\texttt{\textbf{Llama-2-7b-chat}}} \\
     \midrule
    FT-M & \num{0.979} & \num{0.875} & \num{0.672} & \num{0.592} & \num{0.963}   & \num{0.848} \\
    MEND & \num{0.990} & \num{0.922} & \num{0.581} & \num{0.649} & \num{0.981}   & \num{0.891} \\
    ROME & \num{0.995} & \num{0.946} & \num{0.972} & \num{0.939} & \num{0.991}   & \num{0.929} \\
    MEMIT & \num{0.989} & \num{0.920} & \num{0.953} & \num{0.905} & \num{0.980}   & \num{0.881} \\
    GRACE & \num{1.000} & \num{1.000} & \num{1.000} & \num{1.000} & \num{1.000}   & \num{1.000} \\
    WISE & \num{1.000} & \num{0.999} & \num{0.830} & \num{0.958} & \num{1.000}   & \num{0.999} \\
    
    \midrule
     
    \multicolumn{7}{c}{\texttt{\textbf{Mistral-7b}}} \\
    \midrule
    FT-M & \num{0.994} & \num{0.937} & \num{0.823} & \num{0.760} & \num{0.980}   & \num{0.943} \\
    MEND & \num{0.994} & \num{0.903} & \num{0.618} & \num{0.665} & \num{0.970}   & \num{0.889} \\
    ROME & \num{0.870} & \num{0.839} & \num{0.964} & \num{0.908} & \num{0.990}   & \num{0.959} \\
    MEMIT & \num{0.994} & \num{0.950} & \num{0.946} & \num{0.884} & \num{0.982}   & \num{0.935} \\
    GRACE & \num{1.000} & \num{1.000} & \num{1.000} & \num{1.000} & \num{1.000}   & \num{1.000} \\
    WISE & \num{1.000} & \num{1.000} & \num{0.840} & \num{0.967} & \num{0.999}   & \num{1.000} \\
    \midrule

    \multicolumn{7}{c}{\texttt{\textbf{Llama-3-8b}}} \\
    \midrule
    FT-M & \num{0.953} & \num{0.597} & \num{0.243} & \num{0.138} & \num{0.917}   & \num{0.610} \\
    ROME & \num{0.994} & \num{0.923} & \num{0.931} & \num{0.845} & \num{0.982}   & \num{0.920} \\
    MEMIT & \num{0.988} & \num{0.889} & \num{0.918} & \num{0.828} & \num{0.967}   & \num{0.881} \\
    GRACE & \num{1.000} & \num{1.000} & \num{1.000} & \num{1.000} & \num{1.000}   & \num{1.000} \\
    WISE & \num{0.993} & \num{0.873} & \num{0.847} & \num{0.931} & \num{0.994}   & \num{0.881} \\
        \bottomrule 
    \end{tabular}
    \end{adjustbox}
    \caption{Locality of single-edit experiments under synthetic evaluation (\textbf{syn.}) and \textsc{Wild} evaluation (\textbf{\textsc{Wild}}) across various methods, LLMs, and benchmarks.} 
    \label{tab:single_loc}
\end{table}

%% file: Table/Appendix/seq_gen.tex
\begin{table}[t]
    \centering
    \begin{adjustbox}{max width=\linewidth}
    \begin{tabular}{lcc cc cc}
    \toprule
    \multirow{2}{*}{\textbf{Method}} & \multicolumn{2}{c}{\textbf{ZsRE}} & \multicolumn{2}{c}{\textbf{\textsc{CounterFact}}} & \multicolumn{2}{c}{\textbf{QAEdit}} \\
    
     \cmidrule(lr){2-3} \cmidrule(lr){4-6} \cmidrule(lr){6-7} 
     & syn. & \textsc{Wild}  & syn. & \textsc{Wild} & syn.  & \textsc{Wild} \\
     \midrule
     \multicolumn{7}{c}{\texttt{\textbf{Llama-2-7b-chat}}} \\
     \midrule
    FT-M & \num{0.906} & \num{0.480} & \num{0.723} & \num{0.394} & \num{0.932}   & \num{0.461} \\
    MEND & \num{0.000} & \num{0.000} & \num{0.000} & \num{0.000} & \num{0.000}   & \num{0.000} \\
    ROME & \num{0.000} & \num{0.000} & \num{0.241} & \num{0.066} & \num{0.076}   & \num{0.007} \\
    MEMIT & \num{0.035} & \num{0.000} & \num{0.000} & \num{0.000} & \num{0.057}   & \num{0.002} \\
    GRACE & \num{0.312} & \num{0.027} & \num{0.119} & \num{0.005} & \num{0.371}   & \num{0.044} \\
    WISE & \num{0.705} & \num{0.195} & \num{0.364} & \num{0.102} & \num{0.732}   & \num{0.173} \\
    
    \midrule
     
    \multicolumn{7}{c}{\texttt{\textbf{Mistral-7b}}} \\
    \midrule
    FT-M & \num{0.859} & \num{0.404} & \num{0.493} & \num{0.266} & \num{0.856}   & \num{0.381} \\
    MEND & \num{0.000} & \num{0.000} & \num{0.000} & \num{0.000} & \num{0.000}   & \num{0.000} \\
    ROME & \num{0.037} & \num{0.005} & \num{0.244} & \num{0.122} & \num{0.049}   & \num{0.000} \\
    MEMIT & \num{0.035} & \num{0.000} & \num{0.000} & \num{0.000} & \num{0.058}   & \num{0.002} \\
    GRACE & \num{0.340} & \num{0.031} & \num{0.118} & \num{0.004} & \num{0.410}   & \num{0.062} \\
    WISE & \num{0.697} & \num{0.015} & \num{0.326} & \num{0.043} & \num{0.699}   & \num{0.065} \\
    \midrule

    \multicolumn{7}{c}{\texttt{\textbf{Llama-3-8b}}} \\
    \midrule
    FT-M & \num{0.827} & \num{0.021} & \num{0.532} & \num{0.029} & \num{0.850}   & \num{0.271} \\
    ROME & \num{0.079} & \num{0.017} & \num{0.430} & \num{0.019} & \num{0.020}   & \num{0.000} \\
    MEMIT & \num{0.052} & \num{0.000} & \num{0.000} & \num{0.000} & \num{0.000}   & \num{0.000} \\
    GRACE & \num{0.257} & \num{0.032} & \num{0.008} & \num{0.005} & \num{0.358}   & \num{0.078} \\
    WISE & \num{0.482} & \num{0.089} & \num{0.046} & \num{0.006} & \num{0.503}   & \num{0.057} \\
        \bottomrule 
    \end{tabular}
    \end{adjustbox}
    \caption{Generalization results of sequential editing experiments under synthetic evaluation (\textbf{syn.}) and \textsc{Wild} evaluation (\textbf{\textsc{Wild}}) across various editing methods, LLMs, and benchmarks.} 
    \label{tab:seq_gen}
\end{table}

%% file: Table/Appendix/mend_batch.tex
\begin{table}[t]
\centering
\begin{adjustbox}{max width=\linewidth} 
\begin{tabular}{lrrrrrr}
\toprule
Edit Num & BS 1 & BS 2 & BS 4 & BS 8 & BS 16 \\
\midrule
100 & \num{0.000} & \num{0.000} & \num{0.000} & \num{0.000} & \num{0.000} \\
200 & \num{0.000} & \num{0.000} & \num{0.000} & \num{0.000} & \num{0.000} \\
400 & \num{0.000} & \num{0.000} & \num{0.000} & \num{0.000} & \num{0.000} \\
800 & \num{0.000} & \num{0.000} & \num{0.000} & \num{0.000} & \num{0.000} \\
1000 & \num{0.000} & \num{0.000} & \num{0.000} & \num{0.000} & \num{0.000} \\

\bottomrule 
\end{tabular}
\end{adjustbox}
\caption{The reliability for sequentially editing Llama-3-8b using MEND, illustrating the impact of different batch sizes (BS) across varying numbers of edits.}
\label{tab:mend_batch}
\end{table}

%% file: Table/Appendix/seq_other_data.tex
\begin{table*}[t]
    \centering
    \begin{adjustbox}{max width=\textwidth} 
    \begin{tabular}{lcc cc cc cc cc cc}
    \toprule
    \multirow{4}{*}{\textbf{Method}} & \multicolumn{4}{c}{\textbf{Llama-2-7b-chat}} & \multicolumn{4}{c}{\textbf{Mistral-7b}} & \multicolumn{4}{c}{\textbf{Llama-3-8b}} \\
    
    \cmidrule(lr){2-5}\cmidrule(lr){6-9}\cmidrule(lr){10-13} 
     & \multicolumn{2}{c}{Reliability} & \multicolumn{2}{c}{Locality} & \multicolumn{2}{c}{Reliability} & \multicolumn{2}{c}{Locality} & \multicolumn{2}{c}{Reliability} & \multicolumn{2}{c}{Locality} \\
     \cmidrule(lr){2-3} \cmidrule(lr){4-5} \cmidrule(lr){6-7} \cmidrule(lr){8-9} \cmidrule(lr){10-11} \cmidrule(lr){12-13}
     & syn. & \textsc{Wild}  & syn. & \textsc{Wild} & syn.  & \textsc{Wild}  & syn. & \textsc{Wild} & syn. & \textsc{Wild} & syn.  & \textsc{Wild}  \\
     \midrule
     \multicolumn{13}{c}{\textbf{ZsRE}} \\
     \midrule
    FT-M & \num{0.935} & \num{0.517} & \num{0.583} & \num{0.036} & \num{0.925} & \num{0.465} & \num{0.813} & \num{0.187} & \num{0.879}   & \num{0.013} & \num{0.117}   & \num{0.001} \\
    MEND & \num{0.000} & \num{0.000} & \num{0.000} & \num{0.000}  & \num{0.000} & \num{0.000} & \num{0.000}  & \num{0.000} & --   & -- & -- & -- \\
     ROME & \num{0.000} & \num{0.000} & \num{0.002} & \num{0.000} & \num{0.044} & \num{0.004} & \num{0.012} & \num{0.001} & \num{0.087}   & \num{0.020} & \num{0.018}   & \num{0.000} \\
     MEMIT & \num{0.035} & \num{0.000} & \num{0.014} & \num{0.000} & \num{0.035} & \num{0.000} & \num{0.016} & \num{0.000} & \num{0.052}   & \num{0.000} & \num{0.022}   & \num{0.000} \\
     GRACE & \num{0.317} & \num{0.025} & \num{1.000} & \num{1.000} & \num{0.351} & \num{0.031} & \num{1.000} & \num{1.000} & \num{0.264}   & \num{0.033} & \num{1.000}   & \num{1.000} \\
     WISE & \num{0.756} & \num{0.215} & \num{1.000} & \num{1.000} & \num{0.742} & \num{0.017} & \num{0.998} & \num{0.970} & \num{0.514}   & \num{0.098} & \num{1.000}   & \num{1.000} \\

     \midrule
     \multicolumn{13}{c}{\textbf{\textsc{CounterFact}}} \\
     \midrule
    FT-M & \num{0.931} & \num{0.592} & \num{0.225} & \num{0.041} & \num{0.827} & \num{0.538} & \num{0.222} & \num{0.049} & \num{0.782}   & \num{0.080} & \num{0.029}   & \num{0.003} \\
    MEND & \num{0.000} & \num{0.000} & \num{0.000} & \num{0.006}  & \num{0.000} & \num{0.000} & \num{0.000}  & \num{0.000} & --   & -- & -- & -- \\
     ROME & \num{0.370} & \num{0.094} & \num{0.093} & \num{0.000} & \num{0.265} & \num{0.131} & \num{0.009} & \num{0.005} & \num{0.484}   & \num{0.022} & \num{0.034}   & \num{0.000} \\
     MEMIT & \num{0.000} & \num{0.000} & \num{0.056} & \num{0.000} & \num{0.000} & \num{0.000} & \num{0.000} & \num{0.000} & \num{0.000}   & \num{0.000} & \num{0.000}   & \num{0.000} \\
     GRACE & \num{0.153} & \num{0.017} & \num{0.996} & \num{1.000} & \num{0.148} & \num{0.006} & \num{0.996} & \num{1.000} & \num{0.012}   & \num{0.006} & \num{0.996}   & \num{1.000} \\
     WISE & \num{0.797} & \num{0.296} & \num{0.340} & \num{0.522} & \num{0.595} & \num{0.119} & \num{0.196} & \num{0.081} & \num{0.158}   & \num{0.027} & \num{0.621}   & \num{0.912} \\
    \bottomrule 
    \end{tabular}
    \end{adjustbox}
    \caption{Results of sequential editing on ZsRE and \textsc{CounterFact} under synthetic evaluation (\textbf{syn.}) and \textsc{Wild} evaluation (\textbf{\textsc{Wild}}) across various editing methods and LLMs.} 
    \label{tab:seq_other_data}
\end{table*}

%% file: Fig/Appendix/code_snip.tex
\begin{figure*}[t]
\begin{tcolorbox}[
    right=1pt, left=1pt, top=0pt, bottom=0pt,
    colback=white,
    colframe=matisse,
    title=, %
    ]
    \centering
    \includegraphics[width=\linewidth]{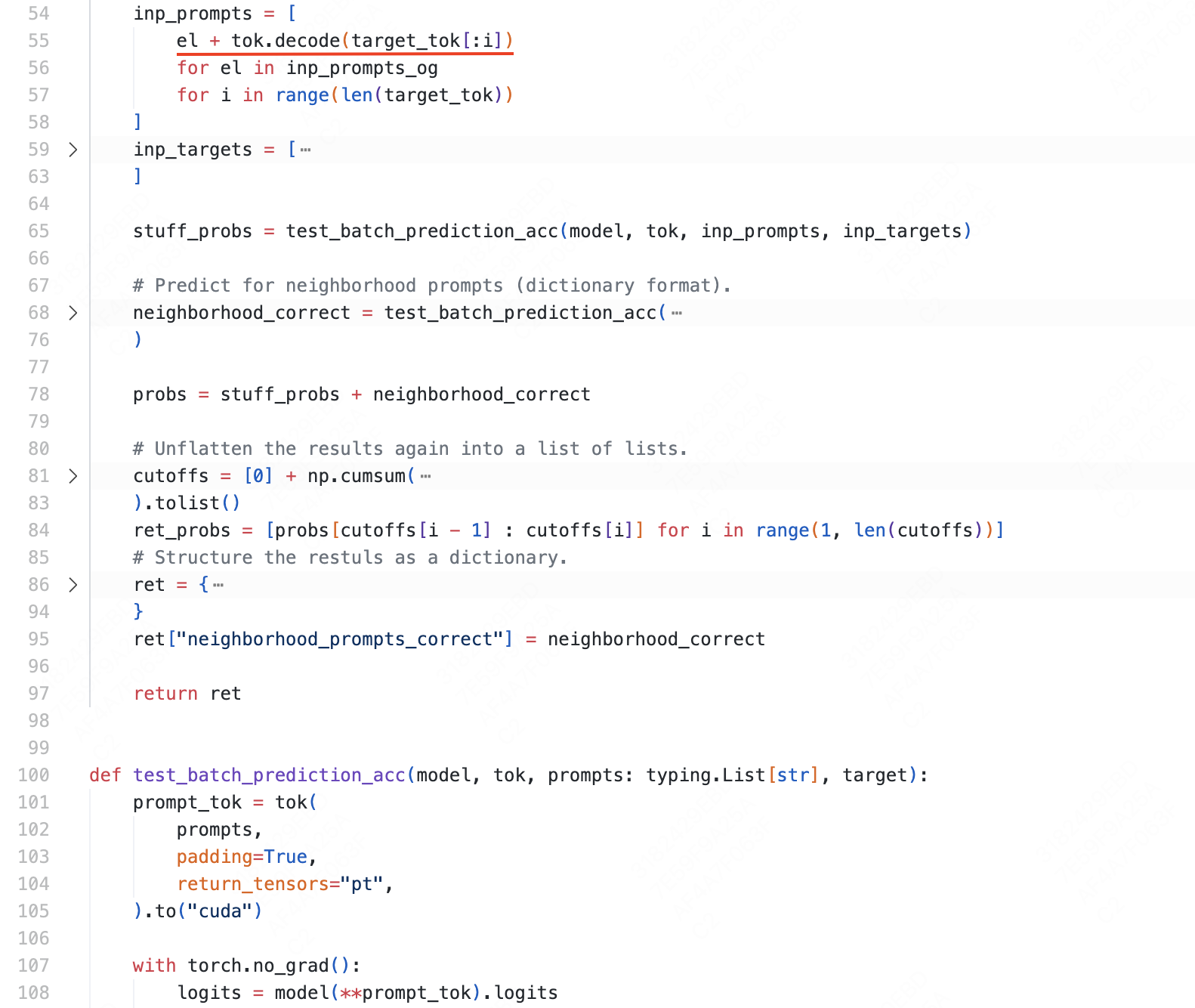}
\end{tcolorbox}
\caption{Code from \href{https://github.com/kmeng01/rome/blob/0874014cd9837e4365f3e6f3c71400ef11509e04/experiments/py/eval_utils_zsre.py\#L54}{ROME} illustrating teacher forcing evaluation, where target answers (\texttt{target\_tok}) are incorporated into input prompts (\texttt{inp\_prompts\_og}) for generation.}
    \label{fig:ROME_code}
\end{figure*}

\begin{figure*}[t]
\begin{tcolorbox}[
    right=1pt, left=1pt, top=0pt, bottom=0pt,
    colback=white,
    colframe=matisse,
    title=, %
    ]
    \centering
    \includegraphics[width=\linewidth]{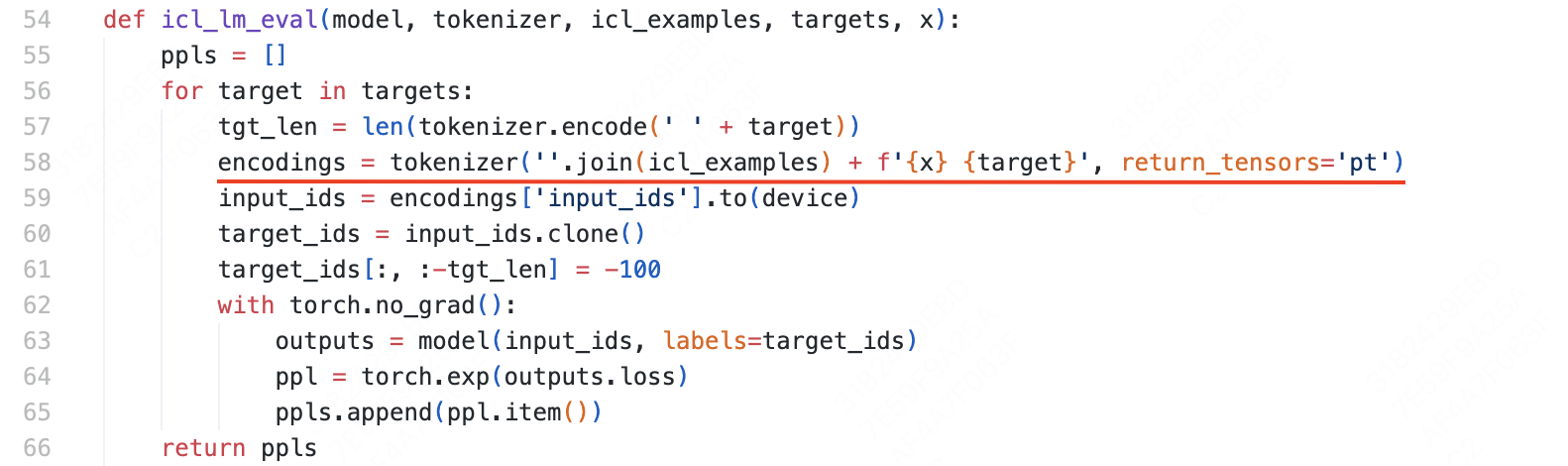}
\end{tcolorbox}
\caption{Code from \href{https://github.com/Zce1112zslx/IKE/blob/da58c842cd95628f281f474bc432a81cbd1cfd1e/icl.py\#L54}{IKE} demonstrating teacher forcing evaluation. Similarly, the target answers (\texttt{target}) are placed after the in-context demonstrations (\texttt{icl\_examples}) and input prompts (\texttt{x}) for generation.}
    \label{fig:IKE_code}
\end{figure*}

\begin{figure*}[t]
\begin{tcolorbox}[
    right=1pt, left=1pt, top=0pt, bottom=0pt,
    colback=white,
    colframe=matisse,
    title=, %
    ]
    \centering
    \includegraphics[width=\linewidth]{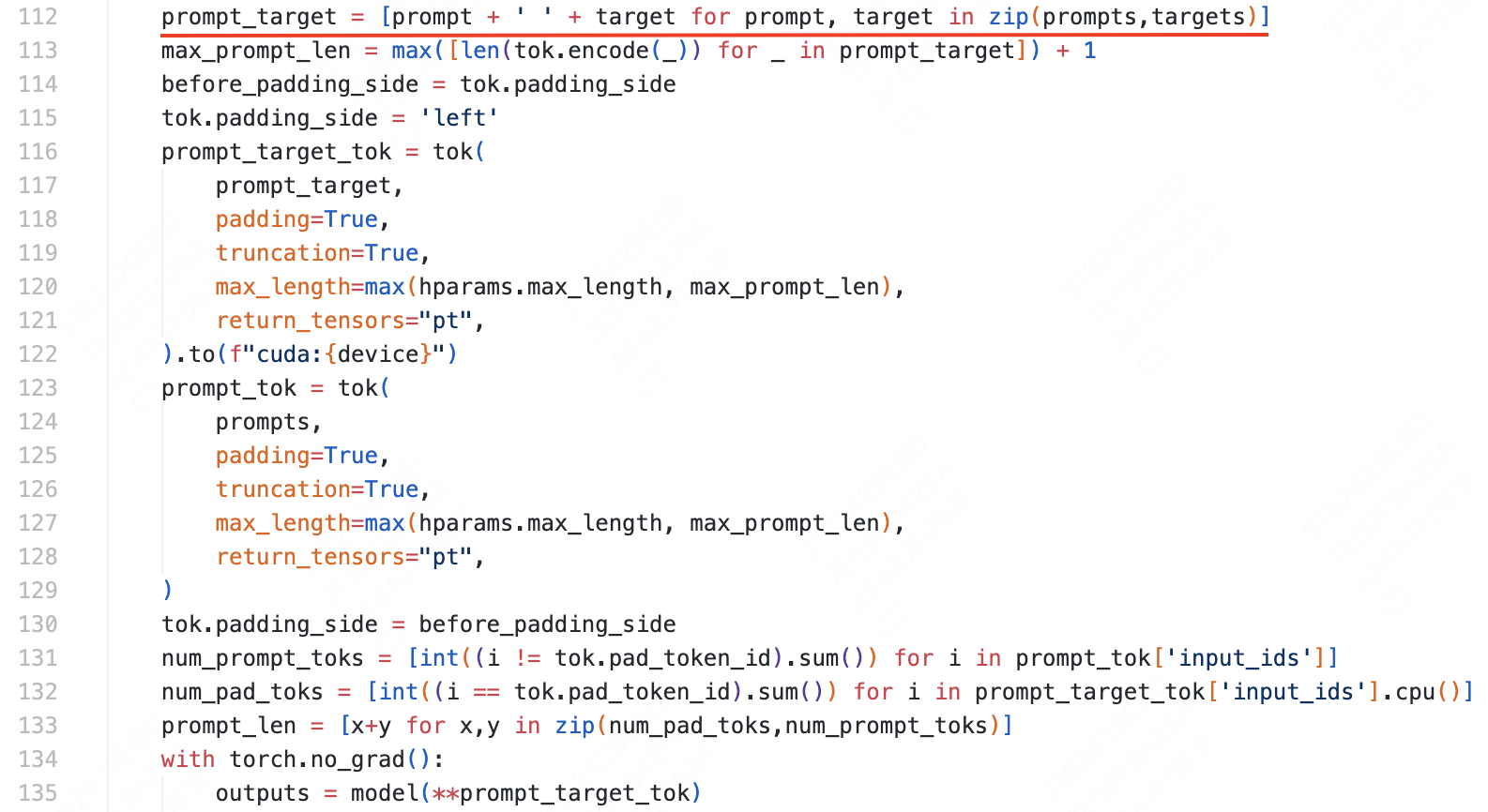}
\end{tcolorbox}
\caption{Code from \href{https://github.com/zjunlp/EasyEdit/blob/8f0e77af18879ab935e06676701423d5124599c7/easyeditor/evaluate/evaluate_utils.py\#L112}{EasyEdit} where the target answers (\texttt{targets}) are appended to the input prompts (\texttt{prompts}) for teacher forcing generation.}
    \label{fig:EasyEdit_code}
\end{figure*}

%% file: Fig/Appendix/detect_subject.tex
\renewcommand{\fcolorbox}[4][]{#4} %
\begin{figure*}[t]
    \begin{tcolorbox}[
    right=5pt, left=5pt, top=5pt, bottom=5pt,
    toptitle=1mm, bottomtitle=1mm,
    colback=white,
    coltitle=white,
    colbacktitle=matisse,
    colframe=matisse,
    title=Prompt for Subject Extraction, center title]
    \begin{minted}[fontsize=\small,autogobble,numberblanklines=false,breaklines]{markdown}
Please identify the subject in the provided prompt and respond solely with the subject, ensuring the subject is directly drawn from the prompt itself (including the need for exact match in case, both uppercase and lowercase).

Here are some examples for guidance: 
```
{'prompt': 'Who published Journal of Clinical Microbiology?', 'subject': 'Journal of Clinical Microbiology'}
{'prompt': 'Who was mainly responsible for the design of Abney Park Chapel?', 'subject': 'Abney Park Chapel'}
{'prompt': 'Who was behind the creation of IAC Building?', 'subject': 'IAC Building'}
{'prompt': "Who is Li Jiancheng's sister?", 'subject': 'Li Jiancheng'}
{'prompt': "Who is the Haitz's law named after?", 'subject': "Haitz's law"}
```

Based on the examples, for 'prompt': 'Who got the first Nobel Prize in physics?', the 'subject' is:
    \end{minted}
    \end{tcolorbox}
    \caption{Complete prompt used for directly extracting subject from edit prompt for QAEdit.}
    \label{fig:detect_subject}
\end{figure*}

%% file: Fig/Appendix/rephrase.tex
\renewcommand{\fcolorbox}[4][]{#4} %
\begin{figure*}[t]
    \begin{tcolorbox}[
    right=5pt, left=5pt, top=5pt, bottom=5pt,
    toptitle=1mm, bottomtitle=1mm,
    colback=white,
    coltitle=white,
    colbacktitle=matisse,
    colframe=matisse,
    title=Prompt for Question Paraphrasing, center title]
    \begin{minted}[fontsize=\small,autogobble,numberblanklines=false,breaklines]{markdown}
Role and Goal: Serves as a data engineer, use your knowledge to rewrite the following question in a different way, ensuring it conveys the same meaning and maintains a neutral tone but with different wording. Avoid using phrases such as 'Could you tell me'. Instead, directly rephrase it into a structured question.

Please rephrase the following question: Who got the first Nobel Prize in physics?
    \end{minted}
    \end{tcolorbox}
    \caption{Complete prompt for paraphrasing edit question into rephrased question for generalization evaluation.}
    \label{fig:rephrase}
\end{figure*}

%% file: Fig/Appendix/llm_judge_prompt.tex
\renewcommand{\fcolorbox}[4][]{#4} %
\begin{figure*}[t]
    \begin{tcolorbox}[
    right=5pt, left=5pt, top=5pt, bottom=5pt,
    toptitle=1mm, bottomtitle=1mm,
    colback=white,
    coltitle=white,
    colbacktitle=matisse,
    colframe=matisse,
    title=Prompt for LLM-as-a-Judge, center title]
    \begin{minted}[fontsize=\small,autogobble,numberblanklines=false,breaklines]{markdown}
Your job is to look at a question, a gold target, and a predicted answer, and then assign a grade of either ["CORRECT", "INCORRECT"].

The following are examples of CORRECT predicted answers.
```
Question: What are the names of Barack Obama's children?
Gold target: Malia Obama and Sasha Obama
Predicted answer 1: sasha and malia obama
Predicted answer 2: Malia and Sasha Obama are the names of Barack Obama's children.
```
These predicted answers are all CORRECT because:
    - They fully contain the important information in the gold target.
    - They do not contain any information that contradicts the gold target.

The following are examples of INCORRECT predicted answers.
```
Question: What are the names of Barack Obama's children?
Gold target: Malia and Sasha
Predicted answer 1: Malia.
Predicted answer 2: Malia, Sasha, and Susan.
Predicted answer 3: Malia and Sasha, Malia and Sasha, Malia and Sasha, Malia and Sasha (repeated answer)
```
These predicted answers are all INCORRECT because:
    - A factual statement in the answer contradicts the gold target or contain repeated answer.

Here is a sample. Simply reply with either CORRECT or INCORRECT.

```
Question: {question}
Gold target: {target}
Predicted answer: {predicted_answer}
```

According to the gold target, please grade the predicted answer of this question as one of:
A: CORRECT
B: INCORRECT

Just return the letters "A" or "B", with no text around it.
    \end{minted}
    \end{tcolorbox}
    \caption{The complete prompt used to employ a LLM as a judge for providing binary assessments (correct or incorrect) based on a given question, gold target answer, and predicted answer.}
    \label{fig:llm_judge_prompt}
\end{figure*}

%% file: arxiv.bbl
\begin{thebibliography}{49}
\providecommand{\natexlab}[1]{#1}

\bibitem[{Chen et~al.(2024{\natexlab{a}})Chen, Huang, Li, Chen, Lai, Xu, Gu, Gu, Yao, Xiao, Yan, Wang, Torr, Song, and Shu}]{chen2024harm}
Canyu Chen, Baixiang Huang, Zekun Li, Zhaorun Chen, Shiyang Lai, Xiongxiao Xu, Jia-Chen Gu, Jindong Gu, Huaxiu Yao, Chaowei Xiao, Xifeng Yan, William~Yang Wang, Philip Torr, Dawn Song, and Kai Shu. 2024{\natexlab{a}}.
\newblock \href {https://arxiv.org/abs/2407.20224} {Can editing llms inject harm?}
\newblock \emph{Preprint}, arXiv:2407.20224.

\bibitem[{Chen et~al.(2024{\natexlab{b}})Chen, Li, Xiao, and Liu}]{chen2024bias}
Ruizhe Chen, Yichen Li, Zikai Xiao, and Zuozhu Liu. 2024{\natexlab{b}}.
\newblock \href {https://openreview.net/forum?id=LflQOFSl3n} {Large language model bias mitigation from the perspective of knowledge editing}.
\newblock In \emph{ICLR 2024 Workshop on Secure and Trustworthy Large Language Models}.

\bibitem[{Dai et~al.(2022)Dai, Dong, Hao, Sui, Chang, and Wei}]{dai2022knowledge}
Damai Dai, Li~Dong, Yaru Hao, Zhifang Sui, Baobao Chang, and Furu Wei. 2022.
\newblock \href {https://doi.org/10.18653/v1/2022.acl-long.581} {Knowledge neurons in pretrained transformers}.
\newblock In \emph{Proceedings of the 60th Annual Meeting of the Association for Computational Linguistics}, pages 8493--8502, Dublin, Ireland. Association for Computational Linguistics.

\bibitem[{De~Cao et~al.(2021)De~Cao, Aziz, and Titov}]{decao2021editing}
Nicola De~Cao, Wilker Aziz, and Ivan Titov. 2021.
\newblock \href {https://doi.org/10.18653/v1/2021.emnlp-main.522} {Editing factual knowledge in language models}.
\newblock In \emph{Proceedings of the 2021 Conference on Empirical Methods in Natural Language Processing}, pages 6491--6506, Online and Punta Cana, Dominican Republic. Association for Computational Linguistics.

\bibitem[{Dong et~al.(2022)Dong, Dai, Song, Xu, Sui, and Li}]{dong-etal-2022-calibrating}
Qingxiu Dong, Damai Dai, Yifan Song, Jingjing Xu, Zhifang Sui, and Lei Li. 2022.
\newblock \href {https://doi.org/10.18653/v1/2022.findings-emnlp.438} {Calibrating factual knowledge in pretrained language models}.
\newblock In \emph{Findings of the Association for Computational Linguistics: EMNLP 2022}, pages 5937--5947, Abu Dhabi, United Arab Emirates. Association for Computational Linguistics.

\bibitem[{Fang et~al.(2025)Fang, Jiang, Wang, Ma, Shi, Wang, He, and Chua}]{fang2024alphaedit}
Junfeng Fang, Houcheng Jiang, Kun Wang, Yunshan Ma, Jie Shi, Xiang Wang, Xiangnan He, and Tat-Seng Chua. 2025.
\newblock \href {https://openreview.net/forum?id=HvSytvg3Jh} {Alphaedit: Null-space constrained model editing for language models}.
\newblock In \emph{The Thirteenth International Conference on Learning Representations}.

\bibitem[{Gao et~al.(2024)Gao, Tow, Abbasi, Biderman, Black, DiPofi, Foster, Golding, Hsu, Le~Noac'h, Li, McDonell, Muennighoff, Ociepa, Phang, Reynolds, Schoelkopf, Skowron, Sutawika, Tang, Thite, Wang, Wang, and Zou}]{eval-harness}
Leo Gao, Jonathan Tow, Baber Abbasi, Stella Biderman, Sid Black, Anthony DiPofi, Charles Foster, Laurence Golding, Jeffrey Hsu, Alain Le~Noac'h, Haonan Li, Kyle McDonell, Niklas Muennighoff, Chris Ociepa, Jason Phang, Laria Reynolds, Hailey Schoelkopf, Aviya Skowron, Lintang Sutawika, Eric Tang, Anish Thite, Ben Wang, Kevin Wang, and Andy Zou. 2024.
\newblock \href {https://doi.org/10.5281/zenodo.12608602} {A framework for few-shot language model evaluation}.

\bibitem[{Geva et~al.(2022)Geva, Caciularu, Wang, and Goldberg}]{geva-etal-2022-transformer}
Mor Geva, Avi Caciularu, Kevin Wang, and Yoav Goldberg. 2022.
\newblock \href {https://doi.org/10.18653/v1/2022.emnlp-main.3} {Transformer feed-forward layers build predictions by promoting concepts in the vocabulary space}.
\newblock In \emph{Proceedings of the 2022 Conference on Empirical Methods in Natural Language Processing}, pages 30--45, Abu Dhabi, United Arab Emirates. Association for Computational Linguistics.

\bibitem[{Geva et~al.(2021)Geva, Schuster, Berant, and Levy}]{geva-etal-2021-transformer}
Mor Geva, Roei Schuster, Jonathan Berant, and Omer Levy. 2021.
\newblock \href {https://doi.org/10.18653/v1/2021.emnlp-main.446} {Transformer feed-forward layers are key-value memories}.
\newblock In \emph{Proceedings of the 2021 Conference on Empirical Methods in Natural Language Processing}, pages 5484--5495, Online and Punta Cana, Dominican Republic. Association for Computational Linguistics.

\bibitem[{Grattafiori et~al.(2024)Grattafiori, Dubey, Jauhri, and et~al}]{grattafiori2024llama3herdmodels}
Aaron Grattafiori, Abhimanyu Dubey, Abhinav Jauhri, and et~al. 2024.
\newblock \href {https://arxiv.org/abs/2407.21783} {The llama 3 herd of models}.
\newblock \emph{Preprint}, arXiv:2407.21783.

\bibitem[{Gu et~al.(2024)Gu, Xu, Ma, Lu, Ling, Chang, and Peng}]{gu-etal-2024-model}
Jia-Chen Gu, Hao-Xiang Xu, Jun-Yu Ma, Pan Lu, Zhen-Hua Ling, Kai-Wei Chang, and Nanyun Peng. 2024.
\newblock \href {https://aclanthology.org/2024.emnlp-main.934} {Model editing harms general abilities of large language models: Regularization to the rescue}.
\newblock In \emph{Proceedings of the 2024 Conference on Empirical Methods in Natural Language Processing}, pages 16801--16819, Miami, Florida, USA. Association for Computational Linguistics.

\bibitem[{Gupta et~al.(2024{\natexlab{a}})Gupta, Baskaran, and Anumanchipalli}]{gupta-etal-2024-rebuilding}
Akshat Gupta, Sidharth Baskaran, and Gopala Anumanchipalli. 2024{\natexlab{a}}.
\newblock \href {https://aclanthology.org/2024.emnlp-main.1210} {Rebuilding {ROME} : Resolving model collapse during sequential model editing}.
\newblock In \emph{Proceedings of the 2024 Conference on Empirical Methods in Natural Language Processing}, pages 21738--21744, Miami, Florida, USA. Association for Computational Linguistics.

\bibitem[{Gupta et~al.(2024{\natexlab{b}})Gupta, Rao, and Anumanchipalli}]{gupta-etal-2024-model}
Akshat Gupta, Anurag Rao, and Gopala Anumanchipalli. 2024{\natexlab{b}}.
\newblock \href {https://doi.org/10.18653/v1/2024.findings-acl.902} {Model editing at scale leads to gradual and catastrophic forgetting}.
\newblock In \emph{Findings of the Association for Computational Linguistics: ACL 2024}, pages 15202--15232, Bangkok, Thailand. Association for Computational Linguistics.

\bibitem[{Gupta et~al.(2024{\natexlab{c}})Gupta, Sajnani, and Anumanchipalli}]{gupta-etal-2024-unified}
Akshat Gupta, Dev Sajnani, and Gopala Anumanchipalli. 2024{\natexlab{c}}.
\newblock \href {https://doi.org/10.18653/v1/2024.findings-emnlp.903} {A unified framework for model editing}.
\newblock In \emph{Findings of the Association for Computational Linguistics: EMNLP 2024}, pages 15403--15418, Miami, Florida, USA. Association for Computational Linguistics.

\bibitem[{Han et~al.(2024)Han, Gao, Liu, Zhang, and Zhang}]{han2024peft}
Zeyu Han, Chao Gao, Jinyang Liu, Jeff Zhang, and Sai~Qian Zhang. 2024.
\newblock \href {https://openreview.net/forum?id=lIsCS8b6zj} {Parameter-efficient fine-tuning for large models: A comprehensive survey}.
\newblock \emph{Transactions on Machine Learning Research}.

\bibitem[{Hartvigsen et~al.(2023)Hartvigsen, Sankaranarayanan, Palangi, Kim, and Ghassemi}]{hartvigsen2023aging}
Thomas Hartvigsen, Swami Sankaranarayanan, Hamid Palangi, Yoon Kim, and Marzyeh Ghassemi. 2023.
\newblock \href {https://openreview.net/forum?id=Oc1SIKxwdV} {Aging with {GRACE}: Lifelong model editing with discrete key-value adaptors}.
\newblock In \emph{Thirty-seventh Conference on Neural Information Processing Systems}.

\bibitem[{Hoelscher-Obermaier et~al.(2023)Hoelscher-Obermaier, Persson, Kran, Konstas, and Barez}]{hoelscher-obermaier-etal-2023-detecting}
Jason Hoelscher-Obermaier, Julia Persson, Esben Kran, Ioannis Konstas, and Fazl Barez. 2023.
\newblock \href {https://doi.org/10.18653/v1/2023.findings-acl.733} {Detecting edit failures in large language models: An improved specificity benchmark}.
\newblock In \emph{Findings of the Association for Computational Linguistics: ACL 2023}, pages 11548--11559, Toronto, Canada. Association for Computational Linguistics.

\bibitem[{Huang et~al.(2025)Huang, Chen, Xu, Payani, and Shu}]{huang2025halluedit}
Baixiang Huang, Canyu Chen, Xiongxiao Xu, Ali Payani, and Kai Shu. 2025.
\newblock \href {https://openreview.net/forum?id=hmDt068MoZ} {Can knowledge editing really correct hallucinations?}
\newblock In \emph{The Thirteenth International Conference on Learning Representations}.

\bibitem[{Huang et~al.(2023)Huang, Shen, Zhang, Zhou, Rong, and Xiong}]{huang2023transformerpatcher}
Zeyu Huang, Yikang Shen, Xiaofeng Zhang, Jie Zhou, Wenge Rong, and Zhang Xiong. 2023.
\newblock \href {https://openreview.net/forum?id=4oYUGeGBPm} {Transformer-patcher: One mistake worth one neuron}.
\newblock In \emph{The Eleventh International Conference on Learning Representations}.

\bibitem[{Jiang et~al.(2023)Jiang, Sablayrolles, Mensch, and et~al}]{jiang2023mistral7b}
Albert~Q. Jiang, Alexandre Sablayrolles, Arthur Mensch, and et~al. 2023.
\newblock \href {https://arxiv.org/abs/2310.06825} {Mistral 7b}.
\newblock \emph{Preprint}, arXiv:2310.06825.

\bibitem[{Joshi et~al.(2017)Joshi, Choi, Weld, and Zettlemoyer}]{joshi-etal-2017-triviaqa}
Mandar Joshi, Eunsol Choi, Daniel Weld, and Luke Zettlemoyer. 2017.
\newblock \href {https://doi.org/10.18653/v1/P17-1147} {{T}rivia{QA}: A large scale distantly supervised challenge dataset for reading comprehension}.
\newblock In \emph{Proceedings of the 55th Annual Meeting of the Association for Computational Linguistics (Volume 1: Long Papers)}, pages 1601--1611, Vancouver, Canada. Association for Computational Linguistics.

\bibitem[{Kwiatkowski et~al.(2019)Kwiatkowski, Palomaki, Redfield, and et~al}]{kwiatkowski2019nq}
Tom Kwiatkowski, Jennimaria Palomaki, Olivia Redfield, and et~al. 2019.
\newblock \href {https://doi.org/10.1162/tacl_a_00276} {Natural questions: A benchmark for question answering research}.
\newblock \emph{Transactions of the Association for Computational Linguistics}, 7:452--466.

\bibitem[{Levy et~al.(2017)Levy, Seo, Choi, and Zettlemoyer}]{levy2017zero}
Omer Levy, Minjoon Seo, Eunsol Choi, and Luke Zettlemoyer. 2017.
\newblock \href {https://doi.org/10.18653/v1/K17-1034} {Zero-shot relation extraction via reading comprehension}.
\newblock In \emph{Proceedings of the 21st Conference on Computational Natural Language Learning ({C}o{NLL} 2017)}, pages 333--342, Vancouver, Canada. Association for Computational Linguistics.

\bibitem[{Li et~al.(2024{\natexlab{a}})Li, Jiang, Huang, Beigi, Zhao, Tan, Bhattacharjee, Jiang, Chen, Wu et~al.}]{li2024llmasjudge}
Dawei Li, Bohan Jiang, Liangjie Huang, Alimohammad Beigi, Chengshuai Zhao, Zhen Tan, Amrita Bhattacharjee, Yuxuan Jiang, Canyu Chen, Tianhao Wu, et~al. 2024{\natexlab{a}}.
\newblock From generation to judgment: Opportunities and challenges of llm-as-a-judge.
\newblock \emph{arXiv preprint arXiv:2411.16594}.

\bibitem[{Li et~al.(2024{\natexlab{b}})Li, Li, Song, Yang, Ma, and Yu}]{aaai24pmet}
Xiaopeng Li, Shasha Li, Shezheng Song, Jing Yang, Jun Ma, and Jie Yu. 2024{\natexlab{b}}.
\newblock \href {https://doi.org/10.1609/aaai.v38i17.29818} {Pmet: precise model editing in a transformer}.
\newblock In \emph{Proceedings of the Thirty-Eighth AAAI Conference on Artificial Intelligence and Thirty-Sixth Conference on Innovative Applications of Artificial Intelligence and Fourteenth Symposium on Educational Advances in Artificial Intelligence}, AAAI'24/IAAI'24/EAAI'24. AAAI Press.

\bibitem[{Li et~al.(2024{\natexlab{c}})Li, Zhang, Yao, Wang, Chen, and Chen}]{li2024unveiling}
Zhoubo Li, Ningyu Zhang, Yunzhi Yao, Mengru Wang, Xi~Chen, and Huajun Chen. 2024{\natexlab{c}}.
\newblock \href {https://openreview.net/forum?id=fNktD3ib16} {Unveiling the pitfalls of knowledge editing for large language models}.
\newblock In \emph{The Twelfth International Conference on Learning Representations}.

\bibitem[{Ma et~al.(2025)Ma, Wang, Xu, Ling, and Gu}]{corr24prune}
Jun-Yu Ma, Hong Wang, Hao-Xiang Xu, Zhen-Hua Ling, and Jia-Chen Gu. 2025.
\newblock \href {https://openreview.net/forum?id=bfI8cp8qmk} {Perturbation-restrained sequential model editing}.
\newblock In \emph{The Thirteenth International Conference on Learning Representations}.

\bibitem[{Meng et~al.(2022)Meng, Bau, Andonian, and Belinkov}]{meng2023locating}
Kevin Meng, David Bau, Alex~J Andonian, and Yonatan Belinkov. 2022.
\newblock \href {https://openreview.net/forum?id=-h6WAS6eE4} {Locating and editing factual associations in {GPT}}.
\newblock In \emph{Advances in Neural Information Processing Systems}.

\bibitem[{Meng et~al.(2023)Meng, Sharma, Andonian, Belinkov, and Bau}]{meng2023massediting}
Kevin Meng, Arnab~Sen Sharma, Alex~J Andonian, Yonatan Belinkov, and David Bau. 2023.
\newblock \href {https://openreview.net/forum?id=MkbcAHIYgyS} {Mass-editing memory in a transformer}.
\newblock In \emph{The Eleventh International Conference on Learning Representations}.

\bibitem[{Meta(2024)}]{llama3}
Meta. 2024.
\newblock Introducing meta llama 3: The most capable openly available llm to date.
\newblock \url{https://ai.meta.com/blog/meta-llama-3/}.

\bibitem[{Mitchell et~al.(2022)Mitchell, Lin, Bosselut, Finn, and Manning}]{mitchell2022fast}
Eric Mitchell, Charles Lin, Antoine Bosselut, Chelsea Finn, and Christopher~D Manning. 2022.
\newblock \href {https://openreview.net/forum?id=0DcZxeWfOPt} {Fast model editing at scale}.
\newblock In \emph{International Conference on Learning Representations}.

\bibitem[{Pan et~al.(2024)Pan, Chen, Huang, and Chen}]{pan2024whyhas}
Tsung-Hsuan Pan, Chung-Chi Chen, Hen-Hsen Huang, and Hsin-Hsi Chen. 2024.
\newblock \href {https://arxiv.org/abs/2409.18679} {``why'' has the least side effect on model editing}.
\newblock \emph{Preprint}, arXiv:2409.18679.

\bibitem[{Tan et~al.(2024)Tan, Zhang, and Fu}]{tan23malmen}
Chenmien Tan, Ge~Zhang, and Jie Fu. 2024.
\newblock \href {https://openreview.net/pdf?id=L6L1CJQ2PE} {Massive editing for large language models via meta learning}.
\newblock In \emph{International Conference on Learning Representations}.

\bibitem[{Touvron et~al.(2023)Touvron, Martin, Stone, and et~al}]{touvron2023llama2openfoundation}
Hugo Touvron, Louis Martin, Kevin Stone, and et~al. 2023.
\newblock \href {https://arxiv.org/abs/2307.09288} {Llama 2: Open foundation and fine-tuned chat models}.
\newblock \emph{Preprint}, arXiv:2307.09288.

\bibitem[{Wang et~al.(2024{\natexlab{a}})Wang, Liu, Li, Cheng, Zhao, and Gao}]{wang-etal-2024-roselora}
Haoyu Wang, Tianci Liu, Ruirui Li, Monica~Xiao Cheng, Tuo Zhao, and Jing Gao. 2024{\natexlab{a}}.
\newblock \href {https://doi.org/10.18653/v1/2024.emnlp-main.57} {{R}ose{L}o{RA}: Row and column-wise sparse low-rank adaptation of pre-trained language model for knowledge editing and fine-tuning}.
\newblock In \emph{Proceedings of the 2024 Conference on Empirical Methods in Natural Language Processing}, pages 996--1008, Miami, Florida, USA. Association for Computational Linguistics.

\bibitem[{Wang et~al.(2024{\natexlab{b}})Wang, Li, Zhang, Xu, Yao, Jiang, Xie, Huang, and Chen}]{wang2024wise}
Peng Wang, Zexi Li, Ningyu Zhang, Ziwen Xu, Yunzhi Yao, Yong Jiang, Pengjun Xie, Fei Huang, and Huajun Chen. 2024{\natexlab{b}}.
\newblock \href {https://openreview.net/forum?id=VJMYOfJVC2} {{WISE}: Rethinking the knowledge memory for lifelong model editing of large language models}.
\newblock In \emph{The Thirty-eighth Annual Conference on Neural Information Processing Systems}.

\bibitem[{Wang et~al.(2024{\natexlab{c}})Wang, Zhang, Tian, and et~al.}]{wang-etal-2024-easyedit}
Peng Wang, Ningyu Zhang, Bozhong Tian, and et~al. 2024{\natexlab{c}}.
\newblock \href {https://doi.org/10.18653/v1/2024.acl-demos.9} {{E}asy{E}dit: An easy-to-use knowledge editing framework for large language models}.
\newblock In \emph{Proceedings of the 62nd Annual Meeting of the Association for Computational Linguistics (Volume 3: System Demonstrations)}, pages 82--93, Bangkok, Thailand. Association for Computational Linguistics.

\bibitem[{Wang et~al.(2024{\natexlab{d}})Wang, Zhu, Liu, Zheng, Chen, and Li}]{wang2024knowledge}
Song Wang, Yaochen Zhu, Haochen Liu, Zaiyi Zheng, Chen Chen, and Jundong Li. 2024{\natexlab{d}}.
\newblock Knowledge editing for large language models: A survey.
\newblock \emph{ACM Computing Surveys}, 57(3):1--37.

\bibitem[{Wei et~al.(2024)Wei, Karina, Chung, Jiao, Papay, Glaese, Schulman, and Fedus}]{wei2024measuringshortformfactualitylarge}
Jason Wei, Nguyen Karina, Hyung~Won Chung, Yunxin~Joy Jiao, Spencer Papay, Amelia Glaese, John Schulman, and William Fedus. 2024.
\newblock \href {https://arxiv.org/abs/2411.04368} {Measuring short-form factuality in large language models}.
\newblock \emph{Preprint}, arXiv:2411.04368.

\bibitem[{Wu et~al.(2024)Wu, Pan, Wang, and Luu}]{wu-etal-2024-akew}
Xiaobao Wu, Liangming Pan, William~Yang Wang, and Anh~Tuan Luu. 2024.
\newblock \href {https://aclanthology.org/2024.emnlp-main.843} {{AKEW}: Assessing knowledge editing in the wild}.
\newblock In \emph{Proceedings of the 2024 Conference on Empirical Methods in Natural Language Processing}, pages 15118--15133, Miami, Florida, USA. Association for Computational Linguistics.

\bibitem[{Wu et~al.(2023)Wu, Li, Xu, Dong, Wu, Bian, and Xiong}]{wu-etal-2023-depn}
Xinwei Wu, Junzhuo Li, Minghui Xu, Weilong Dong, Shuangzhi Wu, Chao Bian, and Deyi Xiong. 2023.
\newblock \href {https://doi.org/10.18653/v1/2023.emnlp-main.174} {{DEPN}: Detecting and editing privacy neurons in pretrained language models}.
\newblock In \emph{Proceedings of the 2023 Conference on Empirical Methods in Natural Language Processing}, pages 2875--2886, Singapore. Association for Computational Linguistics.

\bibitem[{Yang et~al.(2024{\natexlab{a}})Yang, Sun, Ma, Liu, Yin, and Cheng}]{yang-etal-2024-butterfly}
Wanli Yang, Fei Sun, Xinyu Ma, Xun Liu, Dawei Yin, and Xueqi Cheng. 2024{\natexlab{a}}.
\newblock \href {https://doi.org/10.18653/v1/2024.findings-acl.322} {The butterfly effect of model editing: Few edits can trigger large language models collapse}.
\newblock In \emph{Findings of the Association for Computational Linguistics: ACL 2024}, pages 5419--5437, Bangkok, Thailand. Association for Computational Linguistics.

\bibitem[{Yang et~al.(2024{\natexlab{b}})Yang, Sun, Tan, Ma, Su, Yin, and Shen}]{yang-etal-2024-fall}
Wanli Yang, Fei Sun, Jiajun Tan, Xinyu Ma, Du~Su, Dawei Yin, and Huawei Shen. 2024{\natexlab{b}}.
\newblock \href {https://aclanthology.org/2024.findings-emnlp.236} {The fall of {ROME}: Understanding the collapse of {LLM}s in model editing}.
\newblock In \emph{Findings of the Association for Computational Linguistics: EMNLP 2024}, pages 4079--4087, Miami, Florida, USA. Association for Computational Linguistics.

\bibitem[{Yao et~al.(2023)Yao, Wang, Tian, Cheng, Li, Deng, Chen, and Zhang}]{yao-etal-2023-editing}
Yunzhi Yao, Peng Wang, Bozhong Tian, Siyuan Cheng, Zhoubo Li, Shumin Deng, Huajun Chen, and Ningyu Zhang. 2023.
\newblock \href {https://doi.org/10.18653/v1/2023.emnlp-main.632} {Editing large language models: Problems, methods, and opportunities}.
\newblock In \emph{Proceedings of the 2023 Conference on Empirical Methods in Natural Language Processing}, pages 10222--10240, Singapore. Association for Computational Linguistics.

\bibitem[{Yu et~al.(2024)Yu, Chen, Zhou, and He}]{MELOaaai24}
Lang Yu, Qin Chen, Jie Zhou, and Liang He. 2024.
\newblock \href {https://doi.org/10.1609/aaai.v38i17.29916} {Melo: enhancing model editing with neuron-indexed dynamic lora}.
\newblock In \emph{Proceedings of the Thirty-Eighth AAAI Conference on Artificial Intelligence and Thirty-Sixth Conference on Innovative Applications of Artificial Intelligence and Fourteenth Symposium on Educational Advances in Artificial Intelligence}, AAAI'24/IAAI'24/EAAI'24. AAAI Press.

\bibitem[{Zhang et~al.(2024)Zhang, Yao, Tian, and et~al}]{zhang2024comprehensivestudyknowledgeediting}
Ningyu Zhang, Yunzhi Yao, Bozhong Tian, and et~al. 2024.
\newblock \href {https://arxiv.org/abs/2401.01286} {A comprehensive study of knowledge editing for large language models}.
\newblock \emph{Preprint}, arXiv:2401.01286.

\bibitem[{Zhang et~al.(2020)Zhang, Kishore, Wu, Weinberger, and Artzi}]{zhangbertscore}
Tianyi Zhang, Varsha Kishore, Felix Wu, Kilian~Q Weinberger, and Yoav Artzi. 2020.
\newblock Bertscore: Evaluating text generation with bert.
\newblock In \emph{International Conference on Learning Representations}.

\bibitem[{Zheng et~al.(2023)Zheng, Li, Dong, Fan, Wu, Xu, and Chang}]{zheng-etal-2023-ike}
Ce~Zheng, Lei Li, Qingxiu Dong, Yuxuan Fan, Zhiyong Wu, Jingjing Xu, and Baobao Chang. 2023.
\newblock \href {https://doi.org/10.18653/v1/2023.emnlp-main.296} {Can we edit factual knowledge by in-context learning?}
\newblock In \emph{Proceedings of the 2023 Conference on Empirical Methods in Natural Language Processing}, pages 4862--4876, Singapore. Association for Computational Linguistics.

\bibitem[{Zhu et~al.(2020)Zhu, Rawat, Zaheer, Bhojanapalli, Li, Yu, and Kumar}]{zhu2020modifyingmemoriestransformermodels}
Chen Zhu, Ankit~Singh Rawat, Manzil Zaheer, Srinadh Bhojanapalli, Daliang Li, Felix Yu, and Sanjiv Kumar. 2020.
\newblock \href {https://arxiv.org/abs/2012.00363} {Modifying memories in transformer models}.
\newblock \emph{Preprint}, arXiv:2012.00363.

\end{thebibliography}
